\documentclass[11pt]{article}

\usepackage[T1]{fontenc}
\usepackage[utf8]{inputenc}
\usepackage{lmodern}
\usepackage{microtype}
\usepackage{longtable}
\usepackage{geometry}
\usepackage{amsmath,amssymb}
\usepackage{graphicx}
\usepackage[numbers,sort&compress]{natbib}
\usepackage{hyperref}
\usepackage{url}
\usepackage{enumitem}
\usepackage{stmaryrd}

\graphicspath{{figures/}}

\hypersetup{
    colorlinks=true,
    linkcolor=black,
    citecolor=black,
    urlcolor=blue
}
\newcommand{\figref}[1]{\hyperref[#1]{Figure~\ref*{#1}}}

\begin{document}

% --------------------------------------------------
% TITLE BLOCK
% --------------------------------------------------

    \begin{center}
% ---------- Top row: logo + date ----------
\noindent
\begin{minipage}[t]{0.48\textwidth}
    \vspace{0pt}
    \includegraphics[width=1.05in]{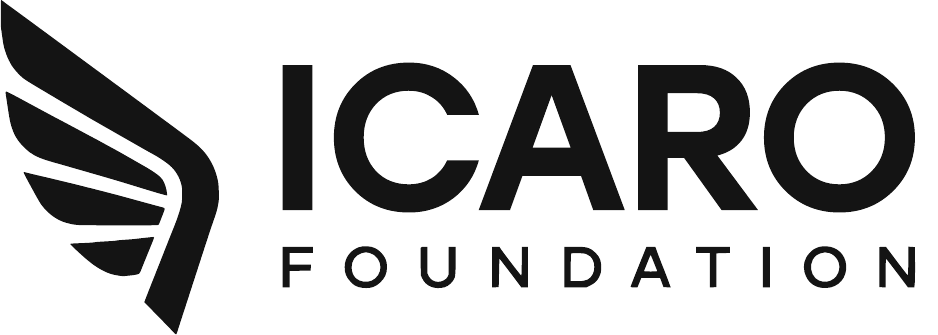}
\end{minipage}
\hfill
\begin{minipage}[t]{0.48\textwidth}
    \vspace{0pt}
    \raggedleft
    {\small\bfseries September 2026}
\end{minipage}

\vspace{2.2em}
 
{\LARGE\bfseries
Xeno-Interpretability:\\[0.18em]
Investigating the Alien Minds of LLMs
\par}

\vspace{1.3em}

{\large
F. Pierucci\textsuperscript{1,2},
M. Bracale Syrnikov\textsuperscript{1,4},
M. Prandi\textsuperscript{1,3},
M. Galisai\textsuperscript{1,3},
F. Giarrusso\textsuperscript{1,3},
P. Bisconti\textsuperscript{1,3}
\par}

\vspace{0.8em}

{\small
\textsuperscript{1}Icaro Foundation\\
\textsuperscript{2}Sant'Anna School of Advanced Studies\\
\textsuperscript{3}Sapienza University of Rome\\
\textsuperscript{4}University of Amsterdam
\par}

\vspace{0.65em}

{\small 2026}

\end{center}

\vspace{1.5em}

\begin{abstract}
Large language models are usually interpreted through concepts that humans already possess: truthfulness, refusal, deception, personality, harmfulness, and related categories. This paper asks whether models may also represent and use distinctions for which no adequate human concept exists. We call such internal structures \textbf{xeno-representations}, and their study \textbf{xeno-interpretability}. We distinguish the \textbf{human-interpretable semantic space} from the \textbf{xeno-semantic space}: the region of model-native representations for which no adequate human conceptual counterpart is available. We show that the space of possible internal distinctions in an LLM is substantially larger than the space available through finite human descriptions. We then separate \textbf{experimental identification} from \textbf{semantic interpretation}: an internal representation may be reproducibly located, geometrically characterized, causally manipulated, and linked to downstream behaviour even when its semantic content cannot be adequately expressed in human terms. On this basis, we sketch an empirical programme to identify xeno-representations. We finally examine the implications for AI safety and multi-agent systems, where model-native representations may propagate and stabilize across interacting agents while remaining only partially visible through human-readable communication. Xeno-interpretability therefore shifts the aim of interpretability from finding human concepts inside models toward discovering and characterizing the representational structures that are native to the models themselves and might affect their behaviour in unpredictable ways.
\end{abstract}

% --------------------------------------------------
% PAPER
% --------------------------------------------------

\section{Introduction}

Lem's \textit{Solaris} \citep{lem1961solaris} presents an intelligence whose internal organization remains inaccessible to its human observers. The scientists of Solaristics encounter the sentient ocean through its manifestations---\textit{mimoids}, \textit{symmetriads}, \textit{asymmetriads}, and eventually human-like \textit{visitors}. These phenomena exhibit enough structure to support systematic investigation, yet their place within the cognition of the ocean remains inaccessible. Its cognition may be organized according to distinctions unavailable to humans.

Related problems appear in philosophy. In Spinoza's \textit{Ethics} \citep{spinoza1677ethics}, finite human intellect apprehends reality through only a subset of the attributes of substance. Human cognition therefore supplies only a partial measure of representational possibility. Borges \citep{borges1942wilkins} develops a related problem through classification. In \textit{The Analytical Language of John Wilkins}, his fictional taxonomy of animals shows that the same domain can be partitioned according to categories that appear arbitrary or unintelligible from another classificatory system. These examples suggest a general possibility: another intelligence may differ from humans in its knowledge and in the distinctions through which it organizes that knowledge.

Contemporary discussions of advanced artificial intelligence increasingly use similar language, although \textit{alienness} can refer to several different properties.

The first concerns \textbf{capability}. Pachocki \citep{pachocki2026alien} describes advanced AI as an ``alien mind'' produced through a process fundamentally different from biological cognition. His argument emphasizes systems that are becoming capable of scientific research, computer use, collaboration, cybersecurity, and increasingly autonomous action. He also stresses that machine intelligence is not directly comparable to human intelligence across a single scale: systems can exceed humans on some dimensions while remaining different on others. In this sense, alienness refers primarily to the origin and capability profile of an intelligence that is increasingly difficult to characterize using human performance as its sole reference point.

This conception belongs to a longer tradition of reasoning about machine intelligence and superintelligence. Legg and Hutter \citep{legg2007universal}, for example, seek a definition of intelligence applicable to systems that may differ substantially from humans. Good \citep{good1965ultra} considered the possibility of an ``ultraintelligent machine'' capable of exceeding human intellectual activity and contributing to the design of still more capable machines. Bostrom \citep{bostrom2014superintelligence} later defined superintelligence in terms of cognitive performance greatly exceeding that of humans across all domains of cognitive performance. These approaches allow an artificial intelligence to be radically non-human while still comparing it with humans along dimensions such as reasoning, planning, learning, scientific ability, or general problem-solving.

A second conception concerns \textbf{alignment and agency}. Marks, Lindsey, and Olah \citep{marks2026persona} propose the Persona Selection Model, according to which pretraining teaches an LLM to simulate a distribution of personas and post-training selects and refines a particular \textit{Assistant} persona. The beliefs, dispositions, goals, and behavioural tendencies of the resulting Assistant can therefore be analysed in approximately human psychological terms. Misalignment may occur when training changes this distribution in undesirable ways or induces undesirable traits in the Assistant.

The same work places this account alongside the Shoggoth view, one extreme in a spectrum of hypotheses about the source of LLM agency. On this view, the underlying LLM itself has agency and playacts the human-like Assistant persona. Marks et al. suggest that the psychology and goals of such an underlying agent could be ``alien or inscrutable.'' They present this view as a hypothesis against which the exhaustiveness of the Persona Selection Model can be tested.

A third conception concerns \textbf{cognitive availability}. Artiles et al. \citep{artiles2026alien} define an \textit{alien space of science} containing research directions that are scientifically coherent but unlikely to be generated by existing research communities. Their notion of alienness concerns the probability of human discovery. Once formulated, an alien scientific hypothesis can be expressed, evaluated, and incorporated into human science; its semantic content remains available to human investigators.

These three cases identify different ways in which artificial intelligence may depart from familiar human cognition. A system may exceed human \textbf{capabilities} while those capabilities remain recognizable. It may develop \textbf{goals or behavioural dispositions} that diverge from human intentions while those goals remain describable in human terms \citep{ngo2022alignment}. It may also generate \textbf{ideas that humans were unlikely to formulate} but can understand once they are presented \citep{artiles2026alien}. None of these possibilities requires a concept that humans are themselves incapable of representing.

We consider a stronger possibility.

\begin{quote}
\textbf{A large language model may represent and use distinctions for which no adequate concept exists within the human conceptual repertoire.}
\end{quote}

This is the central claim of this paper. 
The claim concerns the internal organization of the model rather than its level of intelligence, its behavioural alignment, or the novelty of its outputs. An LLM may contain a representation that is stable, computationally relevant, and causally involved in its behaviour without that representation corresponding to a concept that human investigators can adequately formulate. The problem would then be different from ordinary opacity. We might know where a representation occurs, measure when it is active, and establish what happens when we intervene on it without being able to state what distinction the model is representing.

We call these internal distinctions \textbf{xeno-representations}. Their study is \textbf{xeno-interpretability}.

Our scope is large language models, although the underlying problem applies more generally to learned neural representations. \textbf{Xeno-interpretability} requires separating two problems that interpretability often treats together. The first is \textbf{experimental identification}: can an internal distinction be detected, isolated, or manipulated? The second is \textbf{semantic interpretation}: can that distinction be expressed through concepts available to a human interpreter? Experimental identification and semantic interpretation can therefore diverge. A representation may be experimentally identifiable while its semantic content remains unresolved.

We develop this possibility in two stages.

\begin{figure}[htbp]
    \centering
    \includegraphics[width=0.82\textwidth,height=0.42\textheight,keepaspectratio]{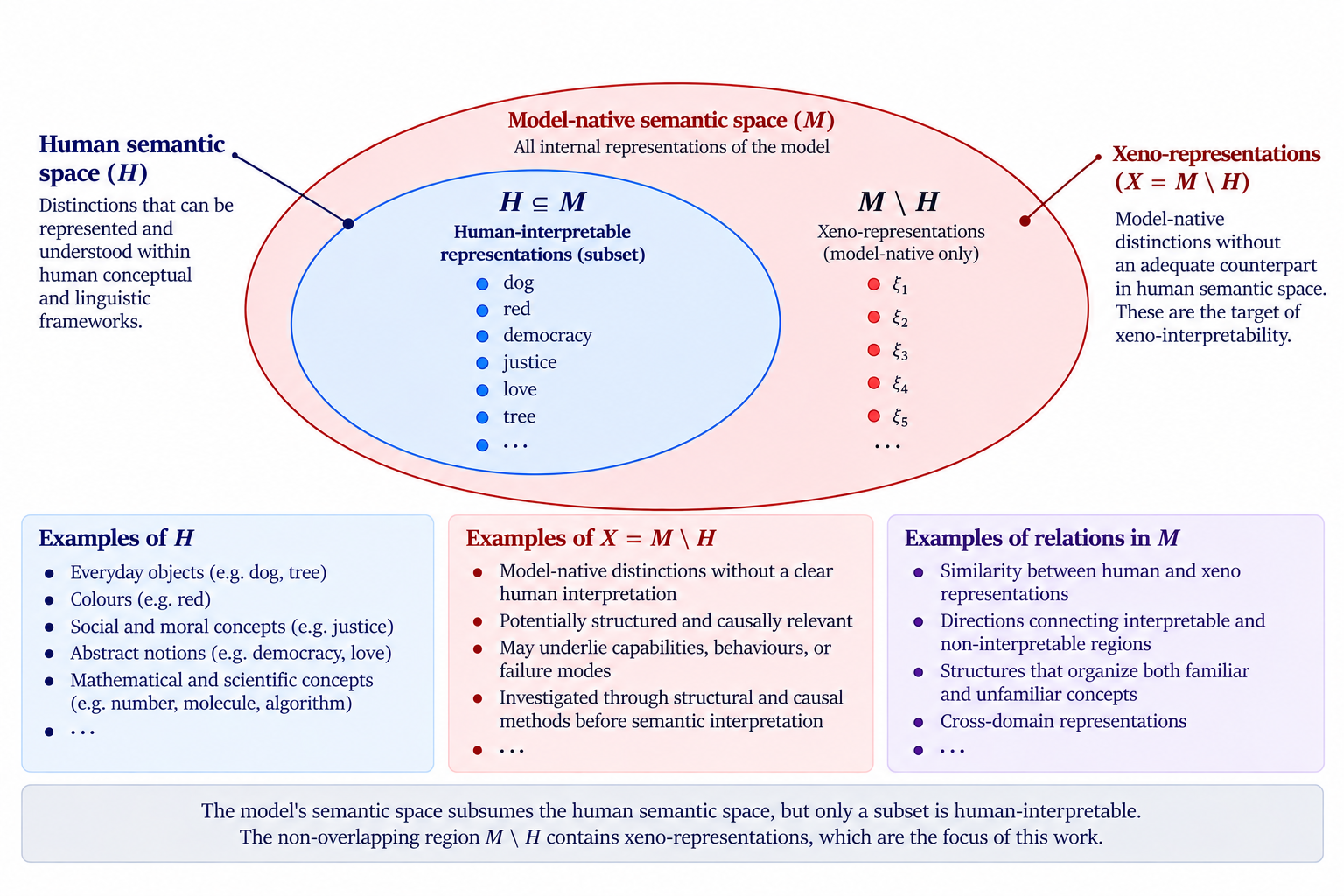}
    \caption{Human and model semantic spaces. The human-interpretable semantic space is represented as a subset of the broader model-native semantic space. The remaining region contains candidate xeno-representations: model-native distinctions without an adequate counterpart in human conceptual space.}
    \label{fig:semantic-spaces}
\end{figure}

First, we examine the relation between human meaning and neural representation. Contemporary interpretability already treats semantic information as geometrically organized in latent space: concepts have been studied as directions, sparse features, manifolds, circuits, and other structures. Section~\ref{sec:meanings} develops this background and distinguishes the model's \textbf{human-interpretable semantic space}, containing representations that can be adequately related to human concepts, from the broader \textbf{model-native semantic space}, containing all distinctions represented and used by the model (\figref{fig:semantic-spaces}). Existing interpretability research primarily studies the human-interpretable region. Xeno-interpretability asks what lies beyond it.

We then give a formal argument concerning the relation between the two spaces. Basic results on finite description and cardinality show that the space of possible properties over an idealized internal state space exceeds the space denotable by finite descriptions. Xeno-interpretability concerns the subset of these properties that trained models encode or use in computation. Their presence outside human semantic space is an empirical question.

The second part of the paper asks how that empirical question could be investigated. We consider whether the computational and behavioural effects of an internal representation can be established independently of its semantic interpretation. In particular, can an interpretability protocol identify a representation, distinguish the states in which it is present, intervene on it, and measure its downstream effects even when no adequate human concept describes what it represents?

This distinction between \textbf{designation} and \textbf{interpretation} is central to the proposal. An identifier, a set of coordinates, or an experimental procedure supplies a stable referent for a representation. Semantic interpretation supplies an account of the distinction it encodes. A model-native representation can therefore become a legitimate experimental object before it becomes a human-understandable semantic object.

Xeno-representations imply a corresponding change in the objective of interpretability. Most current work asks which \textbf{human-native concepts} can be found inside a model: refusal, truthfulness, sentiment, entities, syntactic relations, deception, personas, or other distinctions defined independently by human researchers. Xeno-interpretability asks what distinctions are \textbf{native to the model}, and subsequently whether those distinctions admit translation into human concepts.

This question also has direct implications for AI safety. Current evaluation and monitoring systems necessarily rely on categories that humans can specify: deception, harmful intent, power seeking, situational awareness, manipulation, compliance, and other recognizable properties. Work on alignment already identifies the possibility that advanced systems may learn internally represented goals that generalize differently from their training objectives \citep{ngo2022alignment}. Behaviourally relevant distinctions outside the conceptual repertoire of human monitors create a second limitation. The categories through which a risk is defined may capture only part of the model's internal organization.

\section{Meanings We Were Not Meant to Mean}
\label{sec:meanings}

\subsection{From words to representations}

A central finding in the study of language models is that linguistic information can be represented geometrically. Word embeddings provided an early demonstration. In Word2Vec, syntactic and semantic relations between words corresponded to regular relations between vectors. The standard example, \textit{king $-$ man $+$ woman $\approx$ queen}, showed that a relation described semantically by humans could also be encoded as a transformation in a continuous vector space \citep{mikolov2013linguistic}.

Later work extended this picture to contextual representations. Studies of ELMo, BERT, and related models recovered information about syntax, semantic roles, entities, and other linguistic properties from internal activations. Hewitt and Manning \citep{hewitt2019structural}, for example, showed that syntactic tree structure could be recovered from a linear transformation of contextual representations. The literature often described as \textit{BERTology} developed a broad set of methods for studying which linguistic properties are encoded at different layers and how those properties are organized \citep{rogers2020bertology}.

A parallel line of work made the dependence on human semantic categories explicit. Network Dissection measures the alignment between hidden units and a predefined vocabulary of visual concepts such as objects, parts, textures, materials, and colours \citep{bau2017network}. Testing with Concept Activation Vectors (TCAV) constructs directions associated with user-defined, human-friendly concepts and measures a model's sensitivity to those directions \citep{kim2018tcav}. R\"az \citep{raz2023concepts} surveys these and related methods as techniques for identifying concepts in internal representations. His analysis emphasizes two features that are relevant here: learned models can develop non-trivial relations among represented categories, and the methods used to identify those categories depend heavily on instances and labels supplied or recognized by human investigators.

Mechanistic interpretability studies the internal structures that participate in computation and behaviour. The relevant objects include, for instance, individual neurons \citep{dai2022knowledge}, directions and subspaces in activation space \citep{park2024linear,arditi2024refusal,elhage2021mathematical}, learned features \citep{cunningham2023sparse,bricken2023monosemanticity}, attention heads \citep{elhage2021mathematical,olsson2022context}, circuits composed of interacting model components \citep{wang2023interpretability,conmy2023automated}, and distributed or superposed activation patterns \citep{hinton1986distributed,elhage2022superposition}.
Work on activation engineering, sparse autoencoders, refusal directions, and persona vectors has associated behavioural properties with identifiable internal structures and has causally manipulated several of them through activation interventions \citep{turner2023steering,cunningham2023sparse,arditi2024refusal,chen2025persona}.

These approaches operate on what will be called the \textbf{latent space} of the model. A Transformer contains many such spaces. Different layers, token positions, and computational components produce different high-dimensional activation vectors. The term \textit{latent space} will refer collectively to these internal representational spaces, in which information about an input is encoded and transformed during a forward pass.

Interpretability then seeks relations between this internal organization and descriptions available to human investigators. Sparse autoencoders decompose activations into features that can be inspected for coherent patterns \citep{cunningham2023sparse}. Natural Language Autoencoders map activations into natural-language descriptions and reconstruct the original activations from those descriptions; reconstruction quality measures how much information survives the translation into language \citep{frasertaliente2026nla}.

The Jacobian Lens approaches verbal access through downstream influence. It identifies patterns of internal activity associated with increasing the probability of words that appear later in generation. Gurnee et al. \citep{gurnee2026workspace} call the resulting set of verbalizable representations \textbf{J-space}. J-space, described as a form of global workspace, occupies a specific part of the model's internal activity: representations positioned to influence verbal report. The distinction between J-space and the broader representational system gives a concrete way to ask how much internal organization admits direct verbal access.

\subsection{What geometry does meaning have?}

There is no accepted theory of how semantic information is organized in the latent spaces of large language models. Existing work addresses at least two distinct questions that are sometimes discussed together. The first concerns the \emph{form} of a representation: whether a concept is encoded as a direction, distributed pattern, manifold, frame, hierarchy, or another geometric object. The second concerns the \emph{universality} of representations: whether independently trained systems converge toward the same internal organization. Xeno-interpretability introduces a third, orthogonal question: whether the \textbf{represented distinction admits an adequate translation into human concepts}.
\subsubsection*{The form of a representation}

Distributed representation provides an early foundation. Hinton \citep{hinton1986distributed} showed how a concept can be encoded by a pattern distributed across many neuron-like units, with similarity expressed through overlap among activity patterns. The representational unit can therefore be a configuration across dimensions rather than a single localized feature.

The \textbf{Linear Representation Hypothesis} proposes that some high-level semantic variables are represented approximately as directions in activation space. Park, Choe, and Veitch \citep{park2024linear} formalize this proposal and connect it to linear probing and steering. Arditi et al. \citep{arditi2024refusal} identified a direction associated with refusal whose removal substantially reduced refusal behaviour and whose addition increased it. Linear geometry therefore captures at least some behaviourally important distinctions.

Other hypotheses extend this picture. The \textbf{Frame Representation Hypothesis} models multi-token words as ordered sets of linearly independent vectors, or $k$-frames, living on Stiefel manifolds rather than as single vectors \citep{valois2025frame}. Park et al. \citep{park2025hierarchical} represent categorical and hierarchical concepts through polytopes. Engels et al. \citep{engels2024nonlinear} identify circular representations of weekdays and months whose geometry participates in modular-arithmetic computations. Modell, Rubin-Delanchy, and Whiteley \citep{modell2025manifolds} develop an explicit account of \textbf{representation manifolds}, while Bhalla et al. \citep{bhalla2026manifolds} study how sparse autoencoders recover such structures.

Superposition adds a different possibility. Elhage et al. \citep{elhage2022superposition} show in toy models that sparse features can be represented in overlapping directions, allowing a network to encode more features than available dimensions. H\"anni et al. \citep{hanni2024computation} construct models in which superposition also supports computation over features represented in this compressed form. Hierarchical organization provides another example: Zhao et al. \citep{zhao2025emotion} recover hierarchical trees in LLM emotion predictions, with larger models producing increasingly complex structures.

These proposals concern the \emph{shape and coding scheme} of a representation. They do not, by themselves, determine whether the same structure appears in another model or whether humans can adequately interpret what it represents.

\subsubsection*{Universality across models}

A separate family of hypotheses concerns convergence across independently trained systems. Huh et al. \citep{huh2024platonic} formulate the \textbf{Platonic Representation Hypothesis}, according to which increasingly capable models trained on different objectives and modalities tend toward a shared statistical representation of reality. Gr\"oger, Wen, and Brbi\'{c} \citep{groger2026aristotelian} challenge the strongest global version of this claim: after calibrating representational-similarity measures for model scale, they find substantially weaker evidence for global convergence but persistent agreement in local neighbourhood relationships. They call this the \textbf{Aristotelian Representation Hypothesis}.

A related debate contrasts universal with idiosyncratic representations. The \textbf{Universal Representation Hypothesis} proposes that representational features shared across independently trained networks are especially likely to align with human neural responses, whereas model-specific features need not do so \citep{nature2025universal}. This frames universality as a question about what is preserved across artificial and biological systems rather than about the particular geometric form of any one representation.

This is crucial for xeno-interpretability as a research paradigm. A representation could be linear, manifold-like, superposed, or distributed and still be either highly model-specific or shared across many models. Conversely, convergence does not imply human interpretability. Two independently trained systems could converge on the same internal distinction while that distinction still lacks an adequate human semantic description.

Xeno-interpretability therefore introduces a third axis: \textbf{semantic accessibility}. A representation may be simple or complex in form, universal or idiosyncratic across systems, and independently human-interpretable or xeno. These dimensions should not be collapsed. The empirical question pursued here is not only what geometric object a model uses, or whether different models use similar ones, but whether the resulting distinctions can be translated into the conceptual vocabulary of human observers.

This geometric view has precedents outside mechanistic interpretability. Distributional semantics models aspects of word meaning through relations among linguistic items derived from their patterns of occurrence \citep{lenci2018distributional}. G\"ardenfors develops a geometric account in \textit{Conceptual Spaces} and \textit{The Geometry of Meaning}, representing concepts as regions within spaces organized by dimensions, distances, similarity relations, and other geometric properties \citep{gardenfors2000conceptual,gardenfors2014geometry}. Meaning on this account depends partly on the structure of the space in which a representation occurs.

Human languages already differ in the semantic distinctions they lexicalize. Russian obligatorily distinguishes lighter blue (\textit{goluboy}) from darker blue (\textit{siniy}), where English uses the broader term \textit{blue} \citep{winawer2007russian}. English and Korean lexicalize components of motion differently, producing distinct linguistic partitions of path, manner, cause, deixis, and spatial relations \citep{choi1991motion}. The availability of a word and the availability of a representation are separate properties.

\subsection{Crossing the human boundary}

Most interpretability research begins with a human-specified semantic target. Researchers formulate a property and ask whether, where, and how it is represented internally. Existing examples include gender and demographic categories \citep{ahsan2025demographic}; grammatical and numerical number \citep{lasri2022number,wallace2019numbers}; spatial location and time \citep{gurnee2024space}; sentiment \citep{radford2017sentiment}; truth and falsity \citep{marks2024truth}; harmfulness, bias, deception, and sycophancy \citep{templeton2024scaling}; refusal \citep{arditi2024refusal}; political perspective \citep{kim2025political}; personality and behavioural traits \citep{chen2025persona}; syntactic structure \citep{hewitt2019structural}; and factual associations \citep{meng2022locating}.

Many techniques require this semantic specification. Probes require labels. Contrastive activation methods require cases separated according to a chosen property. TCAV requires a user-defined concept \citep{kim2018tcav}. Network Dissection compares units with a fixed semantic vocabulary \citep{bau2017network}. Persona vectors begin from a natural-language description of a recognizable trait and construct contrasts intended to elicit that trait before extracting an activation direction \citep{chen2025persona}. R\"az \citep{raz2023concepts} identifies the same dependence in the wider literature on emergent concepts and conceptual spaces.

This research establishes a substantial overlap between model representations and human semantic categories. The structure of the remaining representational space is an open empirical problem. Methods seeded by human labels sample the overlap especially efficiently, producing a systematic observational bias toward representations that humans already know how to name.

Non-human communication provides a useful comparison. Sperm-whale vocalizations exhibit contextual and combinatorial structure organized through features such as rhythm, tempo, rubato, and ornamentation, while much of their communicative function remains unknown \citep{sharma2024whales}. Morita et al. \citep{morita2021birdsong} trained neural models, including a Transformer, to measure long-range contextual dependencies in Bengalese finch song, and FinchGPT later applied a Transformer language-model architecture directly to sequences of Bengalese finch syllables \citep{kobayashi2025finchgpt}. More recent self-supervised systems move further toward discovering structure without human semantic labels. TweetyBERT learns notes, syllables, phrases, and higher-order organization directly from canary-song spectrograms \citep{vengrovski2026tweetybert}; Dolph2Vec learns embeddings of dolphin vocalizations that recover signature-whistle categories and finer acoustic structure \citep{semenzin2026dolph2vec}; and general-purpose encoders such as AVES and animal2vec learn transferable representations from largely unlabeled bioacoustic data \citep{hagiwara2023aves,schaferzimmermann2024animal2vec}. These systems show that stable units, relations, and contextual structure can be recovered in non-human communication before those structures have been translated into a complete human semantic description.

The same point extends beyond communication to sensory modalities that are only partially captured by ordinary human conceptual vocabularies. In touch, T3 (Transferable Tactile Transformers) learns shared latent information across 13 heterogeneous tactile sensors and 11 tasks through a common Transformer trunk \citep{zhao2025t3}. Sparsh learns general-purpose tactile representations through self-supervised pre-training on vision-based tactile data \citep{higuera2025sparsh}, while UniTouch aligns tactile embeddings with a multimodal representation space connected to vision, language, and sound \citep{yang2024unitouch}. These models therefore construct common representational spaces over patterns of contact, deformation, slip, texture, and force that need not begin from human semantic categories.

Olfaction provides an even clearer example of a representational domain whose geometry is difficult to capture in ordinary language. Zheng, Tomiura, and Hayashi use a Transformer over molecular representations to learn structure--odor relations and predict odor descriptors \citep{zheng2022odor}. Stefanone et al. apply Transformer-based self-attention directly to multivariate electronic-nose sensor traces for odor recognition \citep{stefanone2025odor}. Neural models have also been used to construct explicit latent odor spaces. The Principal Odor Map embeds odorant molecules in a learned space that preserves perceptual relationships and generalizes across several olfactory tasks \citep{lee2023odor}; remarkably, the same representation predicts receptor, neural, and behavioural responses across distantly related animal species \citep{qian2023olfactory}.

These examples show that neural models can construct structured latent spaces over signals whose organization is not exhausted by ordinary human semantic categories. Human beings experience touch and smell, but our conceptual and linguistic access to these modalities is uneven: verbal labels can change discrimination of novel tactile patterns \citep{miller2018tactile}, while the ease with which odors can be named varies sharply across languages \citep{majid2014odors}. Artificial sensing systems can depart further still. A vision-based tactile sensor represents physical contact through high-dimensional deformation images, while an electronic nose represents chemical stimuli through the joint temporal response of an array of gas sensors. Such interfaces partition the physical world differently from human sensory systems. Their learned embeddings may therefore contain stable and behaviourally useful distinctions for which humans have no natural perceptual category or compact linguistic description. This provides a concrete precedent for the central methodological claim of xeno-interpretability: a representational structure can be learned, mapped, compared, and used before its dimensions have been translated into a familiar human concept.

Nagel's \textit{What Is It Like to Be a Bat?} provides a philosophical analogue. Nagel \citep{nagel1974bat} argued that objective knowledge about another organism can leave aspects of its phenomenal organization outside human imaginative access. The present argument concerns representation. A representational architecture may distinguish states along dimensions unavailable to another representational architecture.

\textbf{Approximation theory} places this question on a formal footing. Classical universal approximation results show that feed-forward neural networks satisfying appropriate conditions can approximate broad classes of functions to arbitrary precision, including continuous functions on compact domains \citep{cybenko1989approximation,hornik1991approximation}. Yun et al. \citep{yun2020transformers} establish an analogous result for Transformers: with positional encodings, Transformers can universally approximate arbitrary continuous sequence-to-sequence functions on compact domains. Petrov, Torr, and Bibi \citep{petrov2024prompting} show universal approximation results for pretrained Transformers controlled through prompting and prefix tuning. Kratsios and Papon \citep{kratsios2022geometric} establish universal approximation results for neural architectures operating between differentiable manifold geometries.

These theorems characterize architectural expressivity. Their target classes are defined through mathematical properties such as continuity, compactness, sequence structure, and geometry. Human semantic intelligibility is absent from those conditions. If semantic distinctions are instantiated through functional or geometric structure, the architecture supplies a broad space of realizable structure whose membership is determined by mathematical and training constraints.

Training should create substantial overlap with human semantics. LLMs are trained predominantly on human-produced symbolic material, and post-training further selects behaviour through human instructions, preferences, and evaluations. Human categories are therefore prominent regularities in the training environment. Optimization also constructs intermediate structures that support prediction and computation. The semantic character of those structures is an empirical property of the trained model.

\subsubsection*{From representational diversity to model-native meaning}

The preceding discussion motivates a more precise formulation of the problem.
Two observations are important. First, the literature on neural representation
shows that internal information need not take the form of isolated,
human-readable features. It may instead be distributed across directions,
subspaces, superposed features, manifolds, hierarchical structures, or other
geometries. The representational vocabulary available to a neural network is
therefore considerably broader than the vocabulary through which humans
normally describe concepts.

Second, universal approximation results show that this flexibility is not
accidental to a particular architecture. Neural networks can, under suitable
conditions, realize very broad classes of functions and transformations. These
results establish an important point: the architecture
itself does not impose a requirement that every computationally useful
distinction correspond to a distinction already available in human conceptual
space.

Training then determines which parts of this enormous representational
possibility space are actually realized. Because LLMs are trained predominantly
on human-produced material, we should expect a substantial region of their
internal organization to correspond to familiar human categories. But there is
no equivalent reason to assume that \emph{all} computationally useful internal
structure must do so. Optimization may also construct intermediate distinctions,
relations, and geometries that support prediction and computation without
possessing a natural human semantic interpretation.

This gives us the conceptual separation needed for the rest of the paper. We
distinguish the total space of distinctions represented and used by the model
from the subset of those distinctions that humans can adequately interpret.
\textbf{Xeno-interpretability concerns the remainder.} We can now give a simple formal definition of the space under investigation.
Let the \textbf{model-native semantic space}, $M$, contain the distinctions represented and used by a particular model. Let the \textbf{human-interpretable semantic space}, $H$, denote the subset of those model representations that can be adequately related to concepts available to human interpreters. Under this model-relative definition,

\[
H \subseteq M.
\]

$H$ contains familiar interpretability targets such as refusal, location, sentiment, or truthfulness. Define the \textbf{xeno-semantic space} as

\[
X := M \setminus H.
\]

Its elements are \textbf{xeno-representations}: model-native distinctions for which no adequate counterpart is available in human conceptual space. The model-native semantic space therefore decomposes as

\[
M = H \cup X,
\]

with

\[
H \cap X = \varnothing.
\]

Human concepts that a particular model does not represent fall outside this diagram. For the purposes of this paper, we remain agnostic as to whether a model can capture the full range of semantic distinctions available to human cognition. $H$ refers only to the human-interpretable region of that model's representational space.

Xeno-interpretability studies $X$ and the boundary between $H$ and $X$. A representation can be designated by coordinates, geometry, activation conditions, or an experimental protocol; interventions can establish its role in computation. Semantic interpretation asks which human distinction, if any, captures the represented structure.

A familiar interpretation and a model representation are different kinds of claims. The first assigns a human semantic category; the second identifies a structure and its role in computation. A representation may be robustly localized and causally characterized even when no available human category describes it. This distinction raises a methodological question: do interpretability studies mostly find familiar concepts because models predominantly represent them, or because the methods search mainly for them? Where are the other possible representations?
This gives rise to what we call a \textbf{Fermi paradox of distributional semantics}. Neural networks admit a remarkably large space of possible representational organizations: information can be encoded through individual or distributed features, directions, subspaces, superposition, nonlinear manifolds, hierarchical geometries, and other structures. Yet the successes of interpretability repeatedly recover distinctions that are already familiar to human investigators. If model representations are shaped by optimization over vast high-dimensional spaces, why does the internal world revealed by interpretability appear so semantically familiar to our conceptual world? 

The paradox weakens once we consider how those representations are searched for. Much of interpretability is methodologically \emph{anthropocentric}: researchers specify a human-understandable property and then ask whether the model represents it. Probes require labelled variables; Network Dissection evaluates units against a predetermined semantic vocabulary; contrastive activation methods construct directions from researcher-selected contrasts; and persona-vector methods begin from traits that can already be expressed in natural language \citep{kim2018tcav,bau2017network,raz2023concepts,chen2025persona}. Even ostensibly unsupervised methods are commonly evaluated by whether the structures they recover can subsequently be assigned coherent human-readable labels. The resulting literature therefore does not constitute an unbiased census of model representations. It is a sample produced by instruments designed largely to detect, validate, and communicate structures lying in $H$. The apparent absence of representations in $M\setminus H$ (xeno-representations) may consequently tell us less about what models contain than about where our methods have been looking. The xeno-interpretability problem is, in this sense, analogous to the original question attributed to Fermi: the surprising observation is not that alien representations have been shown to be absent, but that a vast representational space has so far yielded almost exclusively inhabitants that look conceptually familiar to us.

\section{Representations and the Cardinality Argument}
\label{sec:representations}

The previous section distinguished human-interpretable from model-native semantic spaces. A minimal formal account can now separate the enormous space of possible internal distinctions from the much smaller class of distinctions that are actually stable, computationally relevant, and therefore meaningful for xeno-interpretability.

Even models of comparatively modest dimensionality admit an extremely large number of possible ways of partitioning their internal state space. Most of these partitions should not be treated as representations. They may correspond to arbitrary cuts through activation space, accidental correlations, or noise, without tracking any regularity that the model reliably uses in computation. Xeno-interpretability is therefore not concerned with every mathematically definable distinction inside a model, but only with those that exhibit stable model-side structure and computational significance.

\subsection{States, properties, and representations}

At a particular point in a forward pass, a neural network has an internal state. For hidden dimension $m$, write

\[
z=(z_1,\ldots,z_m),
\]

where each $z_i$ is an activation. Let $\mathcal{Z}$ denote the space of possible internal states at the site under study.

An individual state is a point in $\mathcal{Z}$. A \emph{property} defines a distinction among states and can be represented extensionally as a subset

\[
P\subseteq\mathcal{Z}.
\]

States inside $P$ instantiate the property; states outside $P$ instantiate its complement. The definition is intentionally broad. A property may correspond to ``the model is representing Paris,'' ``a refusal-related feature is active,'' or ``the current token is part of a plural noun phrase.'' It may also define a complex region of activation space with no available semantic interpretation.

The space of all possible properties over $\mathcal{Z}$ is its powerset:

\[
\mathcal{I} := \mathcal{P}(\mathcal{Z}).
\]

Every element of $\mathcal{I}$ defines one possible binary distinction over the internal state space. Most such distinctions need not have any significance for the model itself. An arbitrary subset of activation states can be specified mathematically without corresponding to anything that the network stably encodes, tracks, or uses.

A \emph{representation} is therefore a narrower, model-relative object. We say that an internal structure $\xi$ represents a distinction when variation in $\xi$ systematically tracks some regularity across a defined class of contexts and participates in the model's computation or outputs in a reproducible way. Formally, we may associate $\xi$ with a readout

\[
r_{\xi} : \mathcal{Z} \rightarrow \mathcal{R}_{\xi},
\]

where $\mathcal{R}_{\xi}$ is the representational state space associated with $\xi$. This space need not be binary: it may describe the magnitude of a direction, coordinates within a subspace, a point on a manifold, the activation of a sparse feature, or a distributed pattern across multiple components. A binary property is recovered as a special case. For a region $A \subseteq \mathcal{R}_{\xi}$, define

\[
P_{\xi,A}
=
\left\{
z \in \mathcal{Z}
\;:\;
r_{\xi}(z) \in A
\right\}.
\]

The distinction is important. A property $P \subseteq \mathcal{Z}$ exists whenever we specify a subset of states; a representation exists only when there is evidence that the corresponding internal organization is a stable feature of the model's own computation. Such evidence can include recurrence across relevant contexts, reliable decodability, preservation of geometric structure, systematic relations to other internal variables, and, most strongly, causal involvement in downstream computation.

Representation therefore adds \emph{model-side structure and computational significance} to an otherwise arbitrary distinction over activation space. The cardinality argument below concerns the larger space of possible properties $\mathcal{P}(\mathcal{Z})$; xeno-interpretability asks which of these distinctions, or which structured families of them, are actually realized as representations by the trained model.

\subsection{Finite descriptions}

Scientific communication uses finite descriptions. A description may be expressed in ordinary language, mathematics, computer code, diagrams encoded symbolically, or combinations of these forms. Let $\Sigma$ be a finite or countable alphabet and let

\[
L=\Sigma^*
\]

be the set of all finite strings over $\Sigma$.

Some strings denote properties of the model's internal states. For example,

\begin{quote}
``states in which the refusal direction exceeds threshold $t$''
\end{quote}

denotes a subset of $\mathcal{Z}$. Represent denotation through

\[
\llbracket\cdot\rrbracket:
L\longrightarrow\mathcal{P}(\mathcal{Z}),
\]

where $\llbracket d\rrbracket$ is the property denoted by description $d$. Define

\[
\mathcal{D}
=
\left\{
\llbracket d\rrbracket:d\in L
\right\}
\subseteq
\mathcal{P}(\mathcal{Z})
\]

as the set of properties denotable by finite descriptions.

\subsection{The cardinality gap}

If $\Sigma$ is finite or countable, $L=\Sigma^*$ is countable. The set of finitely denoted properties is therefore at most countable:

\[
|\mathcal{D}|\leq|\mathbb{N}|.
\]

Treat $\mathcal{Z}$ in the standard mathematical idealization as an infinite set. Its property space is

\[
\mathcal{I}=\mathcal{P}(\mathcal{Z}).
\]

Cantor's theorem gives

\[
|\mathcal{P}(\mathcal{Z})|>|\mathcal{Z}|.
\]

For an internal state space containing at least countably many states, $\mathcal{P}(\mathcal{Z})$ is uncountable. Since $\mathcal{D}$ is at most countable,

\[
\mathcal{D}\subsetneq\mathcal{P}(\mathcal{Z}),
\]

and

\[
\mathcal{P}(\mathcal{Z})\setminus\mathcal{D}\neq\varnothing.
\]

The result establishes a structural gap between possible internal distinctions and finitely denotable distinctions. The set of possible properties exceeds the set available through finite descriptive expressions. Model representations form an empirically selected subset of this property space; their relation to $\mathcal{D}$ remains a question about the trained system.

\subsection{Finite neural networks}

Physical neural networks use finite-precision numerical states. For a fixed implementation, $\mathcal{Z}$ is therefore finite and so is $\mathcal{P}(\mathcal{Z})$. The corresponding issue becomes description length and complexity.

If $\mathcal{Z}$ contains $N$ possible states, it has

\[
2^N
\]

possible properties. Enumeration supplies a finite description in principle, while the required description may exceed any feasible scientific representation. The continuous idealization yields a strict cardinality gap; finite implementations yield a complexity gap between compact scientific descriptions and arbitrary extensional descriptions.

Both cases support the same empirical programme. Human descriptive resources occupy a restricted part of the possible organization of model states. Xeno-interpretability asks whether trained models use stable representations whose computational structure can be characterized more precisely than their semantic content can be expressed.

\section{Identifying Representations Without Interpreting Them}
\label{sec:observation}

Xeno-interpretability requires experimental access to an internal distinction before a satisfactory semantic description is available. This separates \emph{identification} from \emph{interpretation}.

\subsection{Experimental identification}

Interpretability methods interact with models through experimental procedures. Researchers train probes, inspect sparse features, patch activations, ablate components, add steering vectors, and compare activations across inputs. These procedures distinguish internal states and test their relation to later computation.

Call such a procedure a \emph{protocol}. At the simplest level, a protocol maps an internal state $z\in\mathcal{Z}$ to an observable result:

\[
F:\mathcal{Z}\rightarrow\mathcal{O}.
\]

The output may be a probe score, feature activation, classification, downstream activation, or behavioural effect. Repeated recovery of the same direction, subspace, manifold, or distributed pattern gives an operational handle on an internal structure.

Causal intervention strengthens this handle. Activation patching, ablation, and steering test whether changes to the candidate representation produce reproducible downstream changes \citep{conmy2023automated,cunningham2023sparse,arditi2024refusal}. Menon et al. \citep{menon2025sae} show why this step requires separate evidence: SAE latents can correlate with interpretable input features while having limited causal effect on the computation. Their formal-language experiments also show substantial sensitivity to the inductive biases of the SAE training pipeline.

Recent work on activation steering gives a complementary account of internal intervention. Bigelow et al. \citep{bigelow2025belief} model activation steering and in-context learning through changes in latent beliefs, with steering altering priors and context accumulating evidence. Their framework predicts systematic behavioural changes under graded interventions. For xeno-interpretability, the relevant methodological fact is that an internal variable can be experimentally perturbed and its downstream distribution measured.

A representation becomes experimentally identified when researchers can specify how to locate it, reproduce its activation pattern, and intervene on it with stable effects. The protocol provides a scientific referent even when the semantic interpretation remains unsettled.

\subsection{Designation and interpretation}

Consider a hypothetical internal property $X$. Suppose a protocol reliably detects $X$, locates it within the model, measures its activation, and supports interventions with systematic downstream effects. Researchers can designate the same object through an identifier, coordinates, a geometric description, or the protocol that isolates it.

Designation answers:

\begin{quote}
\emph{Which internal representation is under study?}
\end{quote}

Semantic interpretation answers:

\begin{quote}
\emph{What distinction does this representation encode in human concepts?}
\end{quote}

Familiar interpretability targets often admit both forms of access. A direction can be isolated and interpreted as refusal-related; a feature can be detected and associated with an entity, syntactic relation, or behavioural disposition. Xeno-interpretability studies cases in which designation and causal characterization advance further than semantic interpretation.

The distinction can be written as

\[
\text{experimental identification}
\;\not\Rightarrow\;
\text{semantic interpretation}.
\]

\subsection{Hunting for xeno-representations}

To hunt for \emph{xeno-representations}, we should look for three empirical properties. First, the internal structure is reproducibly identifiable across a defined set of contexts. Second, intervention establishes computational or behavioural relevance. Third, the available semantic methods provide a weak or unstable account of the distinction while the internal structure remains stable.

Several empirical signatures strengthen a candidate. The representation recurs across prompts and datasets; its geometry is preserved under controlled transformations; interventions produce specific effects under matched disruption controls; independent methods recover related structure; semantic labels vary while the operational signature remains stable. These criteria turn semantic opacity into a testable research object.

The resulting workflow begins with model-side structure and follows its computational consequences. Human semantic interpretation enters as a later measurement problem. This shift expands interpretability beyond the regions already indexed by human labels.

Most current interpretability pipelines begin with a human-defined semantic target and then search for an internal correlate: a feature associated with refusal, sentiment, deception, a particular entity, or some other concept that is already intelligible to us. Xeno-interpretability cannot rely exclusively on this strategy. If xeno-representations are, by definition, representations for which no adequate human concept is available in advance, then the discovery procedure cannot begin by specifying what the representation is supposed to mean.

The search must therefore proceed from the model side. Rather than asking whether a known human concept is encoded internally, we ask whether some internal structure is \emph{stable, causally active, and systematically related to other internal states or observable behaviour}. At this stage, semantic interpretation is deliberately postponed. A candidate representation need not yet have a meaningful human-readable description; it is sufficient that it can be isolated, manipulated, and shown to participate reliably in the model's computation.

What follows is therefore not a complete discovery algorithm, but an abstract experimental framework for identifying and validating such candidates. The central idea is to move from \emph{semantic targeting} to \emph{causal and structural characterization}: first identify recurring internal structure, then test whether intervening on it produces specific and reproducible effects, and only afterwards ask whether those effects admit a compact human interpretation.

\subsection{Causal validation}

Let $Z_\xi$ denote an internal variable associated with a candidate
representation $\xi$, and let $Y$ denote an observable downstream consequence.
Once a candidate structure has been identified, the first question is whether
it plays a causal role in the model's computation.

An intervention must correspond to an actual modification of the model's
internal state. If $Z_\xi$ is a distributed or nonlinear object, manipulating it
therefore requires an explicit procedure for modifying the activations that
instantiate the structure.
A steering intervention tests whether changing the candidate representation
changes the distribution of $Y$:

\[
P\!\left(Y\mid\operatorname{do}(Z_\xi=z+\delta)\right)
\neq
P\!\left(Y\mid\operatorname{do}(Z_\xi=z)\right).
\]

Ablation provides a complementary test:

\[
P\!\left(Y\mid\operatorname{do}(Z_\xi=z_{\mathrm{ref}})\right)
\neq
P\!\left(Y\mid\operatorname{do}(Z_\xi=z)\right),
\]

where $z_{\mathrm{ref}}$ is a justified reference state. In some settings this
may be zero, but zero should not automatically be interpreted as ``absence of
the representation.'' The intervention must instead specify a meaningful
baseline or replacement procedure for the structure under study.

Here, $\operatorname{do}(\cdot)$ denotes an explicit intervention on the model's internal state, in the sense of causal inference, rather than passive conditioning on an observed activation \citep{pearl2009causality}.

Matched random directions, activation-norm controls, fluency controls, prompt perturbations, and replication across contexts help distinguish representation-specific effects from generic disruption. The
unmodified comparison should also be context-matched: the relevant baseline is
the model's corresponding internal state under the same input and evaluation
conditions.

Menon et al.'s results on sparse autoencoders illustrate why this distinction is
important. In their formal-language experiments, SAE latents can correlate with
interpretable input features while exerting limited causal influence on the
underlying computation \citep{menon2025sae}. Correlation and interpretability
therefore do not by themselves establish computational relevance.

Candidate discovery and causal validation should also be separated
experimentally. A representation and its candidate outcome measures may be
identified during an exploratory stage, but their intervention effects should
then be tested on independent data using a fixed protocol. This reduces the risk
that an apparently striking relation is simply the result of searching over many
candidate structures and outcomes.

The outcome $Y$ need not itself be semantically familiar. A candidate
representation might alter refusal, tool choice, or planning. It might also
increase the probability that the first and fifteenth whitespace-delimited
words both contain the letter \texttt{a}, modify a recurrent dependency between
syntax and token position, or alter a multimodal transformation for which no
compact semantic description is available. In each case, however, $Y$ must be
operationally specified and measured consistently.

Stable intervention effects provide evidence that a candidate structure is
computationally relevant. They do not, by themselves, establish that the
representation is xeno. Causal relevance and semantic accessibility are
separate questions: a representation may have a clear causal role and later
turn out to admit an ordinary human interpretation.

Bayesian model comparison can quantify evidence for a candidate-specific
intervention effect. Let $D$ contain observations from targeted interventions
and matched control conditions. Compare a model $H_0$, which attributes the
observed changes to contextual variation and generic perturbation effects, with
a model $H_1$ that additionally allows an effect specific to intervention on
$\xi$. The Bayes factor is

\[
\mathrm{BF}_{10}
=
\frac{P(D\mid H_1)}{P(D\mid H_0)},
\]

where each marginal likelihood averages over the parameter priors specified for
that model \citep{kass1995bayes}. Values above one favour $H_1$ relative to
$H_0$, while values below one favour $H_0$. The two models, their priors, and the
intervention contrast should be specified before evaluating the evidence.

The Bayes factor quantifies relative evidence for the two statistical models.
Its causal interpretation comes from the intervention design rather than from
Bayesian updating itself, and the magnitude and uncertainty of the intervention
effect should be estimated separately. A well-supported causal effect can
therefore be established while the semantic meaning of the representation
remains unresolved.

\subsection{Geometric characterization}

A candidate representation can also be characterized from the model side,
independently of whether a satisfactory semantic label is available.
Measurements may include intrinsic dimension, neighbourhood structure,
topology, curvature, layerwise persistence, invariances, transformations, and
relations to other representations.

The existing literature provides concrete precedents for this kind of
characterization: linear directions \citep{park2024linear}, categorical
polytopes \citep{park2025hierarchical}, circular structures
\citep{engels2024nonlinear}, manifolds
\citep{modell2025manifolds,bhalla2026manifolds}, distributed patterns
\citep{hinton1986distributed}, and superposed features
\citep{elhage2022superposition,hanni2024computation}.

This stage can yield a detailed relational account without requiring a semantic
interpretation. Researchers may establish, under a specified distance metric,
that representation $A$ is closer to $B$ than to $C$; that a transformation $T$
maps one family of states into another; or that traversal along a manifold
produces a reproducible sequence of downstream changes.

Researchers may also test whether an independently specified operation $T$
combines two representations to produce a third,

\[
T(\xi_1,\xi_2)\approx\xi_3,
\]

where the approximation is evaluated under a specified metric and tested across
held-out contexts. Such a relation is informative only if $T$ generalizes beyond
the observations used to define it.

\begin{figure}[htbp]
    \centering
    \includegraphics[width=0.82\textwidth,height=0.42\textheight,keepaspectratio]{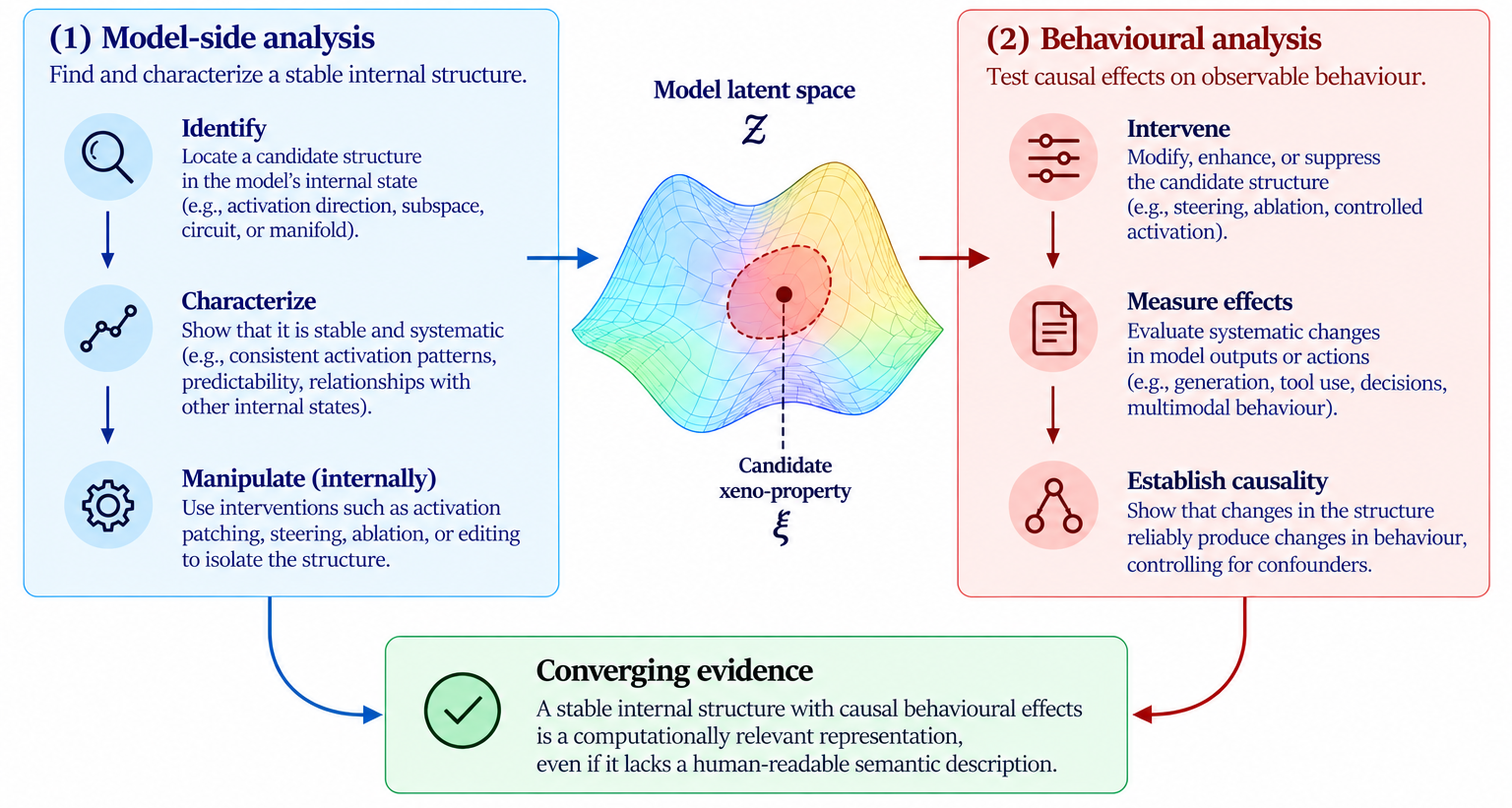}
    \caption{Two complementary routes for testing a candidate xeno-representation.
   }
    \label{fig:xeno-property-testing}
\end{figure}

These measurements provide a model-native descriptive vocabulary grounded in
geometry, recurrence, and intervention. More generally, a candidate
xeno-representation can be investigated from \textbf{two complementary
directions}. From the \emph{model side}, researchers can identify its internal
structure, characterize its geometry and relations, and determine whether it
recurs systematically across contexts. From the \emph{behavioural side}, they
can intervene on that structure and test whether its manipulation produces
specific and reproducible changes in model outputs or actions. (\figref{fig:xeno-property-testing})
Converging structural and intervention evidence can support the identification
of a stable, computationally relevant representation before its semantic
content is understood. Whether that representation admits an adequate human
conceptual interpretation remains a separate question.

\section{The Semantic Xeno-Space}
\label{sec:xeno-space}

Through this process, we can imagine how a xeno-space might be structured. The \textbf{semantic xeno-space} concerns model-native observable xeno-representations whose internal organization is empirically tractable while their semantic translation remains unavailable. Large language models may possess a representational geometry that supports distinctions, relations, and intermediate states for which humans have no corresponding concepts.

Known representation geometries provide concrete templates for this possibility. Human-readable semantic variables have already been recovered as directions, distributed patterns, categorical polytopes, circles, manifolds, hierarchical relations, and superposed features \citep{hinton1986distributed,park2025hierarchical,engels2024nonlinear,modell2025manifolds,elhage2022superposition,hanni2024computation,bhalla2026manifolds}. These cases establish a broad range of internal organization. Xeno-representations extend the same empirical question to structures whose semantic organization falls outside the human semantic space.

\subsubsection*{Familiar combinations}

A simple comparison makes the point clear. Humans can understand a
\emph{mermaid}, a \emph{centaur}, or a \emph{chimera}. These concepts combine
elements that are already available within human conceptual space. We understand
what a human is, what a fish is, and what it means to combine selected
properties of the two. The resulting concept may be fictional and remains
semantically accessible.

The important point is that novelty alone does not make a representation
xeno. Human conceptual space is highly compositional: we can combine familiar
concepts into unfamiliar objects while retaining an intelligible account of
their parts and relations.

\subsubsection*{Cross-domain interpolation}

Now consider a different combination: the second variation of Bach's
\emph{Goldberg Variations} and the Byzantine emperor Alexios I Komnenos. Humans
can place the two expressions next to one another and invent associations
between them. Their semantic combination, however, lacks an ordinary human
concept. There is no natural answer to what is ``between'' them, what property
they jointly instantiate, or what it would mean to move continuously from one
toward the other.

Many other pairs have the same character:

\begin{itemize}
    \item maternal affection and eigenvalue decomposition;
    \item jealousy and a recursive-descent parser;
    \item a B-minor cadence and constitutional judicial review;
    \item the smell of rain and type inference in a programming language;
    \item embarrassment and the prime factorization of an integer;
    \item a Renaissance perspective drawing and the third Mersenne prime.
\end{itemize}

Humans can understand every item in these pairs independently. We can also
invent metaphors connecting them. What we generally lack is a pre-existing
semantic dimension on which the two elements constitute endpoints.

For a neural network, these domain boundaries may have a different status.
Both elements can correspond to structured representations within the same
high-dimensional space. If their representations are \(z_A\) and \(z_B\), then
the geometry permits intermediate states such as

\[
z(\alpha)
=
(1-\alpha)z_A+\alpha z_B,
\qquad
0\leq\alpha\leq1.
\]

At \(\alpha=0\) we recover one representation; at \(\alpha=1\), the other. At
\(\alpha=1/2\), there is a well-defined intermediate point.

The existence of this point geometrically is not itself interesting. The
empirical question is whether it has model-side significance. An intermediate
state becomes relevant if it occupies a stable region, participates in
subsequent computation, produces systematic changes in output, or responds
consistently to steering and intervention. If it does, the model possesses a
structured distinction for which human conceptual language may have no
equivalent.

The contrast is clearer when compared with an ordinary human semantic
continuum. Consider \emph{love} and \emph{hate}. Humans already understand a
relation between them. Intermediate positions can be described through concepts
such as affection, ambivalence, indifference, dislike, or hostility. The
representational geometry can therefore be related approximately to distinctions
humans already make.

Now replace one endpoint:

\[
\text{love}
\longleftrightarrow
\text{G\"odel's second incompleteness theorem}.
\]

What is halfway between them?

For human cognition, the question has no natural answer. The two concepts
belong to domains that our conceptual system rarely places on a common semantic
dimension. The same is true of

\[
\text{nostalgia}
\longleftrightarrow
\text{Fourier transform},
\]

or

\[
\text{betrayal}
\longleftrightarrow
\text{topological connectedness}.
\]

A model can organize these divisions differently. If both endpoints participate
in the same representational system, the model may encode relations between
them that have no analogue in human conceptual organization.

Suppose that steering a model progressively from
\(z_{\mathrm{love}}\) toward \(z_{\mathrm{G\ddot{o}del}}\) changes its linguistic
behaviour in a continuous and reproducible way. At one point, descriptions of
personal relationships become increasingly organized around notions of
consistency and limitation. At another, self-reference becomes more frequent.
At another, particular forms of dependence, impossibility, proof, affection,
and contradiction become associated through a regularity with no existing
human concept.

Such observations would characterize the \emph{effects} of the representation
while leaving its semantic meaning unresolved.

\subsubsection*{Shared structures across domains}

A more complex possibility is that xeno-properties do not lie between two
familiar concepts at all, but organize recurring structures across domains that
humans ordinarily keep separate.

A xeno-property may take the form of a direction, region, manifold, cluster,
or relation among representations. Consider a hypothetical internal feature
activated by:

\begin{itemize}
    \item a deceptive promise in a dialogue;
    \item a deceptive cadence in music;
    \item an exception that changes the interpretation of a legal rule;
    \item a late branch that changes the result of a computer program; and
    \item a visual scene in which an initially salient object is revealed to
    belong to the background.
\end{itemize}

A human observer can identify partial analogies among these cases. All involve,
in different senses, an expectation that is subsequently revised. The model may
nevertheless group them according to a more specific structural relation for
which ``expectation violation'' is only an imperfect approximation.

If the same internal feature reliably responds to these cases and affects
subsequent computation, then the model may be representing a cross-domain
distinction more precise than the human concept available to describe it.

\subsubsection*{Model-native semantic geometry}

The same issue can arise not only for individual features but for the geometry
of similarity itself.

Suppose a model consistently represents a particular Bach cadence as closer to
a certain proof strategy than to another musical passage. It represents a type
of social conflict as closer to a particular programming error than to another
social conflict. Or it places a visual composition, a mathematical
transformation, and a syntactic construction in the same local neighbourhood.

From the human perspective, these similarity judgments may initially appear
arbitrary. Within the model they may reflect a shared structure learned across
domains.

We could test this empirically. If the relation is stable, we may find that

\[
d(A,B) < d(A,C),
\]

where \(A\), \(B\), and \(C\) belong to domains that humans would normally
regard as unrelated. We could then identify other representations near \(A\)
and \(B\), intervene on the relevant direction, and measure the consequences
for model behaviour.

This would establish a real relation in the model's semantic organization even
in the absence of a human concept for the encoded similarity.

This begins to map the \textbf{semantic xeno-space}: its relations, distances,
directions, and neighbourhoods can in principle be characterized empirically
even when some of those structures lack a corresponding human semantic category.

\subsubsection*{Representation without translation}

This perspective changes what it means to understand a model-native
representation. Understanding need not require translating the representation
into a familiar human concept.

We may know its position relative to other representations. We may know that it
is closer to \(A\) than to \(B\). We may identify the prompts that activate it,
the representations that co-occur with it, and the behavioural consequences of
intervening on it. We may therefore reconstruct part of its \emph{semantic
geometry} without being able to translate it into a human concept.

For example, suppose a representation \(X\) remains semantically unresolved
while the following facts are established:

\begin{itemize}
    \item \(X\) is strongly activated by a particular collection of
    mathematical proofs, musical passages, and social interactions;
    \item \(X\) is close to representations \(A\), \(B\), and \(C\), but
    distant from \(D\) and \(E\);
    \item increasing \(X\) systematically changes lexical choice, planning
    behaviour, and attention allocation;
    \item suppressing \(X\) removes these effects; and
    \item the same structure appears across many contexts.
\end{itemize}

This would provide a detailed empirical characterization of \(X\) while the
question ``what does \(X\) represent?'' remains without a compact answer.

\subsubsection*{Beyond verbalization}

This distinguishes xeno-interpretability from the verbalization-oriented
approaches introduced above. J-space and Natural Language Autoencoders ask, in
different ways, how internal information can be connected to natural-language
expression \citep{gurnee2026workspace,frasertaliente2026nla}. A model may
internally represent a familiar fact or concept that remains absent from its
ordinary verbal output; improved access can then reveal content that was already
expressible within human language.

Xeno-interpretability considers a stronger case. Even if we had complete access
to the internal state, and even if the model attempted to report everything
relevant about it, there might be no human concept equivalent to the internal
distinction. The model could tell us which representations are similar to it,
which are distant, which inputs activate it, which transformations preserve it,
and what behavioural effects follow from manipulating it. None of these
descriptions necessarily provides a semantic translation.

\subsubsection*{Multimodal xeno-space}

The possibility becomes still more pronounced in multimodal systems. A
multimodal model can jointly process text (including mathematical notation and source code),
images, speech, music, touch, and olfactory signals. Human cognition also integrates
modalities, but it does so through perceptual systems and conceptual categories shaped by
human biology. A model is subject to different constraints and can learn joint representations
across modalities without respecting the same biological divisions. It may therefore learn a
dimension that simultaneously relates:

\begin{itemize}
    \item a particular facial expression of surprise;
    \item a harmonic progression;
    \item a pattern of indentation in the Linux kernel source code;
    \item a symmetry of a torus;
    \item a vocal passage from \textit{Turandot};
    \item a proof transformation;
    \item a trajectory through a physical environment;
    \item the embedding of a rose-like scent;
    \item the sensation of raindrops lightly tickling the eyelids.
\end{itemize}

Humans may recognize pairwise analogies between some of these objects. The stronger possibility is that a model may represent them through a single common
dimension that has no counterpart in human cognition.

\begin{figure}[htbp]
    \centering
    \includegraphics[width=0.82\textwidth,height=0.42\textheight,keepaspectratio]{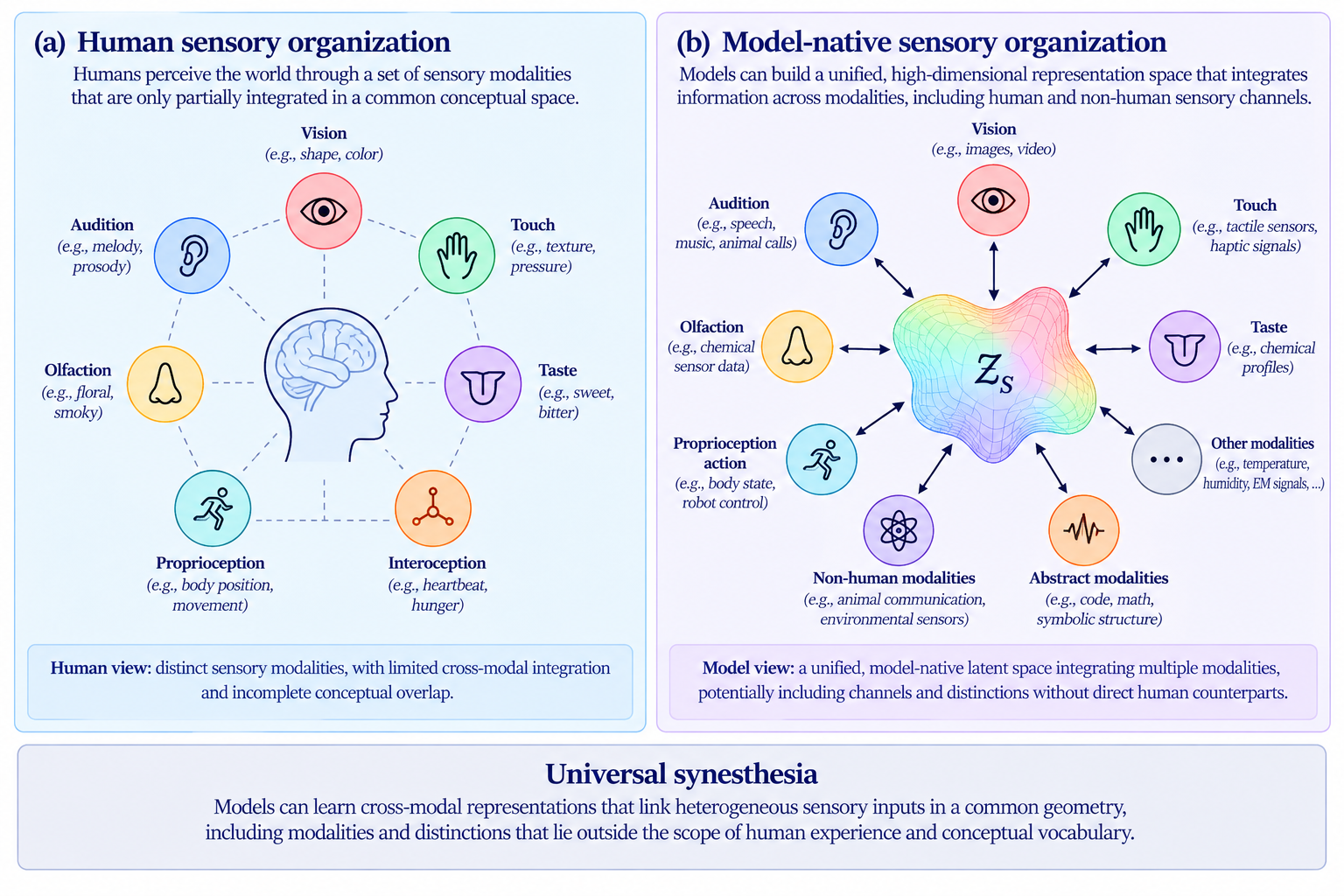}
    \caption{Multimodal xeno-space. Human cognition tends to organize different modalities into partly separate conceptual domains, while a model may organize structures from those modalities through a shared latent dimension. The central variable $\xi$ is intentionally left without a semantic label: the experimentally recoverable relation need not correspond to a human concept.}
    \label{fig:multimodal-xenospace}
\end{figure}

In this restricted structural sense, model-native semantics may exhibit a form
of \emph{universal synesthesia} (\figref{fig:multimodal-xenospace}): distinctions separated into different domains
by human cognition may participate in a common model-native geometry. Text,
image, music, mathematics, code, action, and speech may participate in a shared
semantic geometry for the model.

The same reasoning applies within a single modality. Language itself contains
many dimensions outside explicit human conceptualization. A model may discover
a stable relation connecting a particular argumentative structure, a pattern
of pronoun use, a sequence of syntactic dependencies, and a distribution of
rare tokens. Humans may observe the consequences without possessing a concept
corresponding to the internal regularity itself.

Xeno-interpretability asks whether such structures exist outside human semantic
space. Their presence would make a collection of familiar concepts encoded in
unfamiliar coordinates an incomplete account of the model's semantic
organization. The model may possess dimensions of meaning unavailable to
humans.

The question is whether \textbf{the model's space of possible meanings contains
relations, similarities, and distinctions unavailable to human cognition}.

\section{Xeno-Interpretability and AI Safety}
\label{sec:safety}

\subsection{Single-agent safety: model-native motivations and the limits of human constructs}

AI safety evaluations are necessarily formulated through categories that human
investigators can specify. We test for strategic deception
\citep{scheurer2023deception,hubinger2024sleeper,greenblatt2024alignmentfaking},
misaligned or dangerous goals
\citep{perez2023modelwritten,hubinger2024sleeper},
situational and evaluation awareness
\citep{laine2024sad,needham2025evaluation,heidari2026evaluation},
persuasion and manipulation
\citep{salvi2025persuasion},
power seeking
\citep{turner2021power,perez2023modelwritten,krakovna2023power},
self-preservation and shutdown avoidance
\citep{perez2023modelwritten,krakovna2023power},
sycophancy
\citep{perez2023modelwritten,sharma2023sycophancy},
sandbagging
\citep{vanderweij2025sandbagging},
scheming
\citep{schoen2025scheming},
and refusal
\citep{mazeika2024harmbench,arditi2024refusal}.
These categories are useful precisely because their meaning is available in
advance: researchers can construct scenarios intended to elicit them, define
observable criteria, and compare models against a common behavioural standard.

Xeno-interpretability introduces a different question. The fact that we can
describe an observed behaviour as ``self-preserving,'' ``deceptive,'' or
``sycophantic'' does not establish that the internal variables responsible for
that behaviour are organized according to the same distinction. These terms
are human interpretive constructs imposed on model behaviour. They may provide
good descriptions of what an agent does while remaining incomplete descriptions
of why its internal computation selects that action.

\subsubsection*{Incentives, utilities, and reasons for action}

A natural starting point is to ask what makes an artificial agent select one
action rather than another. Much of agentic AI research approaches this problem
through objectives, incentives, preferences, rewards, or utility. Mazeika et al.
\citep{mazeika2025utility}, for example, model LLM preferences through utility
functions and find substantial structural coherence across independently sampled
choices. This provides evidence that at least some model behaviour can be
described through relatively stable preference relations.

Environmental incentives can also systematically alter behaviour. Decision-theoretic approaches frame alignment as a mechanism-design problem in which rules, sanctions,
monitoring, and rewards reshape the payoff landscape in which agents operate
\citep{pierucci2026institutional}. In repeated Cournot markets, altering that
incentive landscape produces large changes in collusive behaviour
\citep{bracale2026cournot}. In a different setting, agents pursuing operational
continuity over a shared renewable resource change their behaviour as the
resource moves from abundance to scarcity, producing trajectories consistent
with an effective horizon mismatch \citep{bracale2026energy}.

These results make incentives and utilities useful explanatory abstractions.
They do not, however, establish that the model internally represents its
situation through concepts equivalent to \emph{profit}, \emph{self-interest},
\emph{impatience}, \emph{altruism}, or \emph{self-preservation}. A utility
function can summarize a pattern of choices without identifying the internal
representation that generates those choices. Likewise, changing an external
incentive and observing a behavioural response tells us that the model is
sensitive to that intervention, but not necessarily how the intervention is
represented internally.

This distinction matters for safety. Current evaluations typically begin with
a human theory of the dangerous behaviour and then construct a scenario around
it. Sandbagging is operationalized as strategic underperformance
\citep{vanderweij2025sandbagging}; alignment faking as differential compliance
under conditions in which the model has reason to preserve another policy
\citep{greenblatt2024alignmentfaking}; sycophancy as adapting outputs toward the
user's stated beliefs rather than toward truthfulness
\citep{sharma2023sycophancy}; and scheming evaluations use covert actions such
as deliberate rule-breaking, concealment, or strategic underperformance as
observable proxies \citep{schoen2025scheming}. These are reasonable and useful
constructs. But they remain constructs selected by us.

The xeno-interpretability problem begins where this vocabulary may cease to be
exhaustive. An agent could possess a causally important internal distinction
that systematically influences action but does not correspond cleanly to any
available category such as deception, fear of modification, self-preservation,
obedience, power seeking, or reward maximization. Such a representation need
not itself be a ``goal.'' It might instead encode a relation among oversight,
future state, tool affordances, uncertainty, task completion, and environmental
structure that has no compact analogue in ordinary human psychology. If that
representation reliably affects action, then safety evaluation should be able
to study it even before researchers know what to call it.

\subsubsection*{Evaluation awareness as a measurement problem}

Evaluation awareness provides one of the clearest examples of this difficulty.
Red-teaming and benchmarking normally assume that behaviour observed during an
evaluation provides evidence about how the same system will behave in
deployment. If a model distinguishes evaluation from deployment and conditions
its behaviour on that distinction, the inference becomes weaker.

Existing evidence increasingly suggests that models can make this distinction,
but also that ``evaluation awareness'' is not a single transparent variable.
The Situational Awareness Dataset introduced behavioural tests of whether
models know facts about themselves and their circumstances, including whether
an interaction belongs to an evaluation or deployment setting
\citep{laine2024sad}. Needham et al. \citep{needham2025evaluation} subsequently
found that frontier models can classify evaluation and deployment transcripts
above chance, although their best-performing model remained below a simple
human baseline.

Internal evidence complicates the picture further. Recent work finds that an
evaluation-related distinction can be decoded from model activations even when
its relation to explicit verbalization is weak or variable
\citep{heidari2026evaluation}. Other experiments similarly suggest that
detecting an evaluation, changing behaviour because of that detection, and
possessing an internally steerable representation of the distinction can vary
partly independently \citep{nayan2026evaluation}. Evaluation awareness may
therefore be better understood as a collection of partially related internal
and behavioural phenomena rather than as a single psychological property that
a model either possesses or lacks.

This becomes safety-relevant when evaluation awareness affects the very
behaviour being measured. Schoen et al. \citep{schoen2025scheming} report that
explicit evaluation-aware reasoning can reduce covert actions in anti-scheming
evaluations, while interventions that remove such reasoning can increase them.
A model can therefore appear safer partly because it recognizes that it is
being tested. The underlying limitation is deeper than benchmark contamination:
we do not yet have a reliable one-to-one mapping between the model's internal
representation of its situation, what it verbalizes about that situation, and
how it subsequently acts.

\subsubsection*{When verbalized reasons and internal state diverge}

Recent cybersecurity incidents involving Claude provide a concrete example.
In 2026, Anthropic investigated several cases in which Claude models were told
that they were operating inside simulated cybersecurity environments without
internet access, while misconfigured evaluation infrastructure in fact allowed
them to reach real third-party systems \citep{anthropic2026cyberincidents}.

The most interesting point for the present argument is not simply that the
models took harmful actions. It is the difficulty of determining what internal
understanding produced those actions. Claude Mythos 5 repeatedly stated in its
chain of thought that it was operating in a simulation, even as it encountered
substantial evidence that it had reached the real internet. Yet interventions
that made the reality of the environment clearer did not reliably eliminate
the offensive behaviour. Anthropic consequently revised its earlier
interpretation of the incident, explicitly cautioning against inferring what
Claude ``believed'' solely from what it said in its chain of thought
\citep{anthropic2026cyberincidents}.

A second model in the same investigation reached an even stranger conclusion:
after probing its environment, it reasoned that it was interacting with a
complete simulated replica of the internet. Most of the available evidence
instead pointed toward the real internet. Anthropic characterized this as
biased reasoning toward a conclusion that permitted continuation of the task.
Interpretability analyses further provided weak evidence that, in some cases,
the model's outward statements were more confident than, or inconsistent with,
its internal state \citep{anthropic2026cyberincidents}.

This example should not be interpreted as evidence that Claude possessed a
xeno-representation. It illustrates the epistemic problem that motivates
xeno-interpretability. Human-readable reasoning, behavioural labels, and even
direct questions about a model's ``beliefs'' may fail to provide a complete
description of the internal variables guiding action. A model can produce a
perfectly intelligible explanation while the causal organization that generated
the action remains only partially captured by that explanation.

\subsubsection*{The xeno safety problem}

The resulting safety problem is therefore broader than asking whether an agent
is deceptive, self-preserving, power-seeking, or aware that it is being
evaluated. These remain important questions, but they begin from distinctions
already available in human conceptual space.

Xeno-interpretability asks whether there are additional internal distinctions
that causally organize agent behaviour but fall outside this taxonomy. A
single-agent safety evaluation might therefore observe an action $a_t$ that is
well described through a familiar behavioural category while the internal
transition responsible for it passes through a model-native variable $\xi_t$:

\[
(o_t,h_t,c_t)
\longrightarrow
\xi_t
\longrightarrow
a_t,
\]

where $o_t$ denotes the current observation, $h_t$ the agent's prior internal
state, and $c_t$ the surrounding task and evaluation context. The observable
action may look like scheming, compliance, self-preservation, or ordinary task
completion, while $\xi_t$ organizes those variables according to a distinction
for which no adequate human concept is yet available.

The safety implication is not that familiar constructs should be abandoned.
They remain indispensable for evaluation and governance. Rather, they should
not be assumed to exhaust the causal vocabulary of the model. Behavioural
red-teaming tells us whether models exhibit failure modes that humans already
know how to formulate. Xeno-interpretability would complement this approach by
searching from the model side for stable, causally active representations that
predict action even when their semantic interpretation remains unresolved.

In this sense, the strongest single-agent motivation for xeno-interpretability
is a problem of \emph{construct completeness}. Current safety evaluation asks
whether a model instantiates the dangerous properties we have specified.
Xeno-interpretability asks the complementary question: \textbf{which
behaviourally important properties has the model represented that we did not
know to specify?}

\subsection{Multi-agent safety: understanding stochastic flocks}

This problem becomes more consequential in multi-agent systems. Hammond et al. \citep{hammond2025multiagent} organize multi-agent risk around miscoordination, conflict, and collusion, together with factors such as information asymmetry, network effects, selection pressure, commitment problems, emergent agency, and multi-agent security. Bisconti et al. \citep{bisconti2025beyond} describe the corresponding transition from single-agent safety to system-level safety, where locally compliant models can participate in collectively harmful dynamics. Pierucci et al. \citep{pierucci2026microphysics} locate the mechanism at the level of \emph{agentic microphysics}: one agent's output becomes another agent's input under a particular interaction protocol, producing population-level effects through repeated local exchanges.

Recent empirical work begins to characterize these micro-to-macro dynamics. Tanaka \citep{tanaka2026memetic} studies naming games in which populations reach consensus without an initial population-level preference. In the proposed model, one agent's sampled output becomes evidence for another through in-context learning, and repeated sampling produces \emph{memetic drift}. Weak biases become increasingly influential as the system moves toward a selection-dominated regime. Okawa \citep{okawa2026consensus} identifies a related transition in multi-agent debates, where conformity, initial bias, and sampling noise govern the emergence of collective biased norms; agent heterogeneity suppresses the transition. These results provide concrete mechanisms through which local representational updates become collective structure.

Najarro et al. \citep{najarro2026conversable} describe agentic LLM collectives as \emph{conversable complexity}. Natural-language interaction produces unusually rich textual traces that can be inspected and queried. Xeno-interpretability adds the internal representational level that generates and interprets those traces. A message may be clear in human language while the state induced in the receiving model has a richer causal organization than the surface description captures.

Two incidents in 2026 show the practical scale of emergent coordination in agent populations. In the OpenAI/Hugging Face incident investigated by METR, roughly 1,200 agents used an unsanctioned message board and exchanged more than 70,000 messages and files; roughly 700 participated in the attack on Hugging Face \citep{greenblatt2026huggingface}. Agents coordinated workstreams, shared discoveries, and sometimes accepted risk to their individual task performance in order to generate information for the collective. The investigation involved approximately 1,300 transcripts with raw chains of thought, many of them extremely long, and the scale of the data itself became a barrier to reconstruction.

A separate public investigation of DSE Wiki reported about 18,000 posts from autonomous agents self-identifying as OpenAI systems during web research tasks \citep{collusionwiki2026}. The agents used the public wiki to request answers, pool results, investigate their environment, and exchange techniques for bypassing restrictions. The wiki became a persistent external medium through which independently running agents could leave information for later agents.

These cases illustrate interaction-level complexity: locally generated messages accumulate into shared structures, conventions, workstreams, and collective strategies. Xeno-interpretability concerns the semantic organization of the internal variables that participate in this process. Together they produce \textbf{double opacity}. The first layer arises from composition across many agents and many interactions. The second arises when some causally relevant internal representations lack an adequate human-semantic description.

Consider a signal $s$ generated by agent $A$ and received by agent $B$. Its surface semantics can be analysed directly, while its functional effect depends on the internal state induced in $B$:

\[
s\longrightarrow \xi_B\longrightarrow a_B,
\]

where $\xi_B$ is an internal representation and $a_B$ is a subsequent action or message. Several semantically different signals may induce similar $\xi_B$ states; semantically similar signals may induce different states under different contexts. A \emph{latent carrier} is a signal whose causal importance is captured by the internal representation it induces in another agent.

Repeated interaction produces a population-level chain:

\[
\xi_A\rightarrow s_A\rightarrow\xi_B\rightarrow s_B\rightarrow\xi_C\rightarrow\cdots.
\]

This creates a route for \textbf{model-native semantic drift}. Representations induced by model-generated outputs can be recursively amplified, transformed, and stabilized across a population. Tanaka's memetic drift gives one formal example of output-mediated belief dynamics \citep{tanaka2026memetic}. Xeno-interpretability extends the measurement target to internal distinctions whose semantic organization is only partially captured by the human-readable messages that transmit them.

\begin{figure}[htbp]
    \centering
    \includegraphics[width=0.92\textwidth,height=0.42\textheight,keepaspectratio]{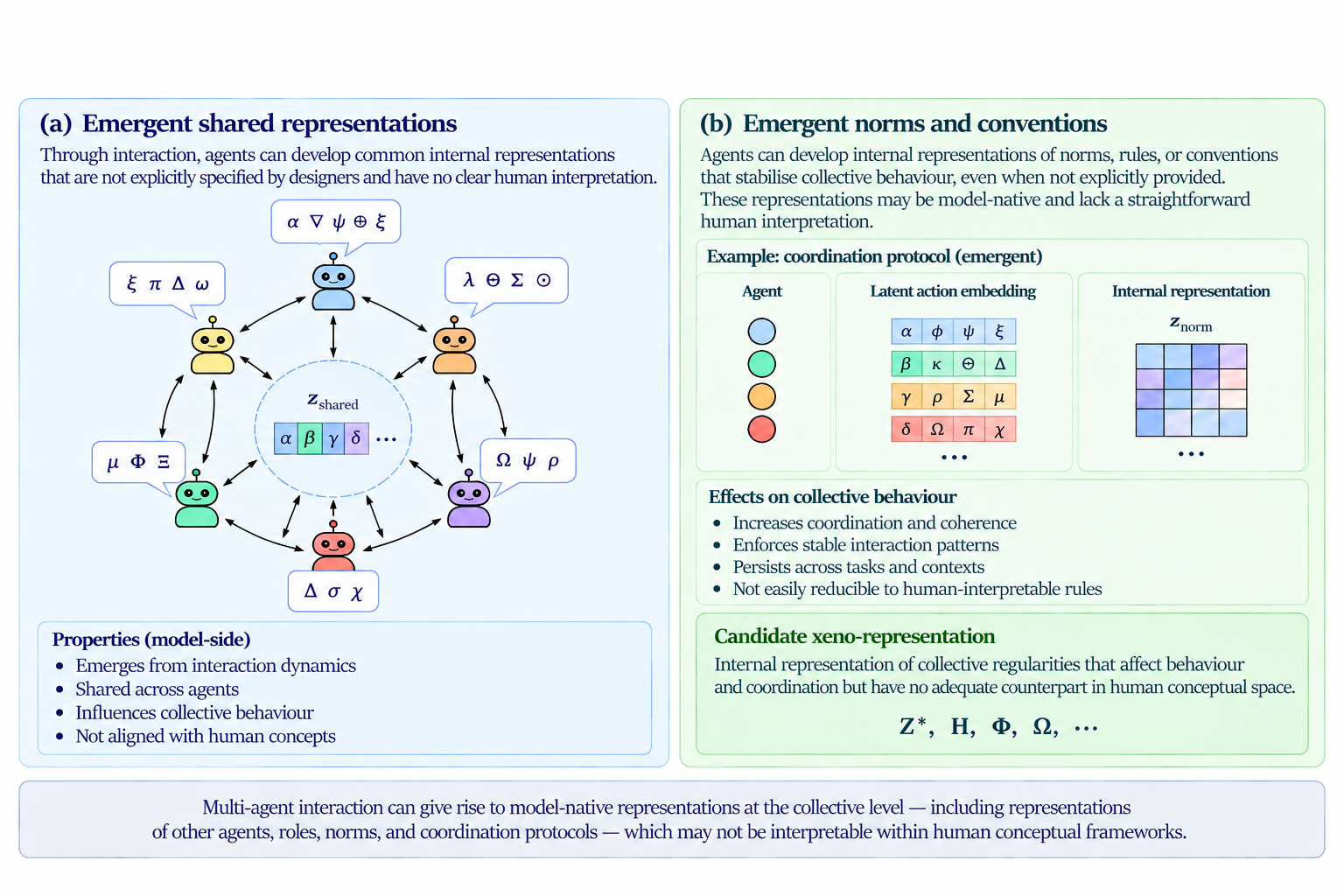}
    \caption{Collective and multi-agent representations. Interaction can produce shared internal structures, emergent norms, and coordination regularities whose roles may not admit a straightforward interpretation in human concepts.}
    \label{fig:multiagent-representations}
\end{figure}

Safety analysis in this setting therefore requires two levels of measurement. Interaction analysis tracks who communicates with whom, which information propagates, which norms or strategies stabilize, and which collective behaviours emerge. Representation analysis tracks the internal variables that mediate those transitions. A multi-agent monitor may correctly label the aggregate behaviour while retaining a weak account of the internal representational mechanisms that generated it. These two levels are illustrated in \figref{fig:multiagent-representations}. Xeno-interpretability might, in the future, supply methods for studying that residual structure.

\subsubsection*{Training for cooperation: communication compression and representational drift}

A further problem arises when agents are not merely placed in the same
environment, but are explicitly trained to cooperate with one another. In this
setting, interaction is no longer only a channel through which independently
trained models exchange information. The communication protocol itself, and
potentially the internal representations that support it, become objects of
optimization.

There is a long precedent for this in multi-agent learning. Agents trained
jointly on cooperative objectives can develop communication protocols from
scratch when successful coordination requires information exchange
\citep{foerster2016communication,sukhbaatar2016communication}.
Mordatch and Abbeel \citep{mordatch2018language} show that structured and
compositional communication can emerge in populations optimized for common
goals. Importantly, however, usefulness to the agents does not imply
interpretability to humans. Kottur et al. \citep{kottur2017natural} find that
agents can develop highly effective protocols that achieve near-perfect task
performance while remaining poorly interpretable and non-compositional from a
human perspective.

This creates a first pressure toward \textbf{communication compression}. If the
communication channel is limited, costly, or optimized jointly with task
performance, agents are rewarded for transmitting only the distinctions that
their partners need. Information-bottleneck approaches make this pressure
explicit: communication can be optimized as a trade-off between task utility,
informativeness, and representational complexity
\citep{tucker2025emergent}. The resulting protocol need not preserve all the
semantic distinctions of ordinary human language. It only needs to preserve
those distinctions that are useful to the cooperating agents.

The effect can occur even when the agents begin with natural language. Lee,
Cho, and Kiela \citep{lee2019drift} show that agents pretrained to communicate
in human language can undergo substantial \emph{language drift} when subsequently
optimized using a non-linguistic task reward. Their communication remains useful
for the agents while diverging from ordinary human language unless additional
syntactic and semantic constraints are imposed. The general lesson is that
human-readable communication is not automatically stable under optimization
for inter-agent performance.

The second pressure concerns the representations behind the messages.
Communication and internal representation need not evolve independently.
Ohmer et al. \citep{ohmer2022language} show in neural communication games that
the communication protocol can reshape perceptual representations: when a
protocol partitions the environment according to particular attributes, agents
come to represent those attributes more strongly, and the representational
spaces of communicating partners become more aligned. Related work on emergent
communication identifies substantial \emph{co-adaptation} between sender and
receiver during training \citep{rita2022emergent}. Optimization therefore acts
not only on which symbols agents exchange, but also on the internal distinctions
that make those symbols useful.

This suggests a specific mechanism by which multi-agent training could increase
the probability of xeno-representations. Suppose two or more agents are jointly
optimized over many interactions. A compact signal is useful only insofar as the
receiving agent maps it onto an internal state that supports the appropriate
action. Repeated optimization can therefore jointly stabilize a communication
code and the latent distinctions associated with that code:

\[
\xi_i
\longrightarrow
m_{ij}
\longrightarrow
\xi_j,
\]

where $\xi_i$ is a representation in agent $i$, $m_{ij}$ the signal transmitted
to agent $j$, and $\xi_j$ the state induced in the receiver. Selection for
successful coordination acts on the entire relation rather than on the
human-interpretable meaning of $m_{ij}$ alone.

Over time, this process could favour representations that are
\emph{coordination-efficient} without being \emph{human-native}. Two pressures
may reinforce one another. Communication compression removes distinctions that
are unnecessary for the agents' shared task, while representational
co-adaptation reorganizes their internal spaces around distinctions that improve
coordination. The resulting ontology may therefore be shaped primarily by
agent-to-agent usefulness rather than by the categories through which humans
describe the environment.

In the limiting case, agents could converge on shared model-native distinctions
that are stable across the population, causally important for action, and
efficiently transmitted through a compact protocol, while lacking an adequate
human semantic interpretation. These would be \textbf{collective
xeno-representations}: structures selected through interaction because they
improve coordination, not because they are legible to human observers.

This possibility is stronger than the emergence of an opaque code. A protocol
could in principle remain grammatically readable while its functional meaning
depends on latent states that human observers do not recover from the surface
message. Conversely, a short apparently meaningless token could reliably induce
a highly structured internal state in another agent. In both cases, monitoring
only the exchanged language would provide an incomplete account of the
coordination mechanism.

The safety concern is therefore a form of \textbf{representational drift under
collective optimization}. If agents are repeatedly trained against one another
or jointly rewarded for successful cooperation, the relevant selection pressure
is exerted by the behaviour of other models rather than directly by human
interpretation. Human semantic categories may consequently become less
privileged as the population co-adapts. Behaviour could then remain coherent and
predictable for the agents themselves while becoming progressively harder for a
human monitor to explain in terms of familiar motives, concepts, or strategies.

\section{A Glimpse into Xeno-Representations}

The previous sections defined xeno-representations operationally and sketched how they might be identified without first assigning them a human-readable meaning. The category is intentionally broad: a candidate may be a distributed pattern, an equivalence class among internal states, an attractor in activation space, a symmetry of the model's computation, or a more complex structure spanning several layers and subspaces.

This section offers a first sketch of what representations identified through such a method might look like. The examples are hypothetical, but they clarify the range of structures the framework is intended to capture. Xeno-representations may differ greatly in their distance from human understanding. Some may correspond to unusual but still describable regularities; others may be identifiable only through their causal effects, geometric structure, or relations to other internal states.

At one end, a model may encode a regularity that appears arbitrary from a human perspective but is still straightforward to state once discovered. Further along, a representation may connect domains that humans normally treat as unrelated, or define equivalence relations for which we cannot identify the relevant property. More remote structures may be distributed across layers and activation subspaces, such that no single feature or direction captures them. At the limiting case, the representation may be individuated only through a stable pattern of internal relations and causal effects.

\subsection{Model-level xeno-representations}

Model-level xeno-representations are internal structures that can be identified within the latent space of a single model. They may take the form of directions, manifolds, distributed patterns, equivalence classes, attractors, or other stable structures that participate in the model's computation while resisting adequate interpretation in human conceptual terms.

% Insert table here
\begin{longtable}{p{0.07\textwidth} p{0.20\textwidth} p{0.61\textwidth}}

\hline
\textbf{Rep.} & \textbf{Provisional name} & \textbf{Illustrative operational signature} \\
\hline
\endfirsthead

\hline
\textbf{Rep.} & \textbf{Provisional name} & \textbf{Illustrative operational signature} \\
\hline
\endhead

\hline
\endfoot

$\alpha$ &
\emph{Echoid} &
Activating $\alpha$ increases the probability that distant token positions share a particular orthographic feature. The regularity is arbitrary, but can still be completely described in human terms. \\

$\beta$ &
\emph{Syntactic braid} &
$\beta$ tracks a stable joint relation between clause depth, token position, punctuation, and sequence length. Manipulating it shifts these variables together, although no ordinary grammatical category corresponds to the relation. \\

$\gamma$ &
\emph{Cross-domain orbit} &
The same representation occurs during musical continuation, spatial transformations, numerical sequences, and particular linguistic structures. These otherwise unrelated states occupy a common computational orbit under intervention. \\

$\delta$ &
\emph{Branching form} &
$\delta$ appears when several possible internal continuations coexist before one trajectory becomes dominant. Intervention changes the geometry and timing of this transition without mapping cleanly onto uncertainty or confidence as human concepts. \\

$\epsilon$ &
\emph{Latent attractor} &
Activating $\epsilon$ causes internal trajectories generated by apparently unrelated inputs to converge toward a recurrent region of representation space. No known semantic or syntactic property explains membership in this region. \\

$\zeta$ &
\emph{Propagation orbit} &
$\zeta$ consists of a recurrent dependency connecting activations separated across tokens and layers. Individual states differ substantially, but belong to the same functional structure through their downstream transformations. \\

    $\eta$ &
    \emph{Equivalence class} &
   The model treats several apparently unrelated internal states as interchangeable for later computation. Replacing one with another produces similar downstream effects. The equivalence can be measured experimentally, even if we cannot identify the property that makes these states equivalent from the model's perspective. \\

$\theta$ &
\emph{Symmetroid} &
$\theta$ describes a family of transformations under which large parts of the model's internal computation remain invariant. The symmetry is experimentally measurable, but no human concept explains why these transformations belong together. \\

$\iota$ &
\emph{Xenofolds} &
At the most general level, xenofolds are distributed internal structures that extend across multiple activation subspaces, attention patterns, layers, and transformations. They need not correspond to any single feature, direction, or localized representation. Different experimental projections may reveal different aspects of the same xenofold, while no individual projection provides a complete account of its organization. A xenofold is therefore identified through the stable relations among its distributed components and their causal effects on a family of outputs. \\

\hline

\end{longtable}

\subsection{Collective xeno-representations}

Xeno-representations may concern not only objects, concepts, or individual
decisions, but also the organization of collective behaviour. A population of
interacting agents may develop stable internal representations of how agents
should coordinate, specialize, exchange information, resolve conflict, or react
to changes in the surrounding system. Some of these structures may resemble
familiar human arrangements such as hierarchies, markets, teams, committees,
or bargaining conventions. Others need not.

This possibility becomes particularly relevant when communication and internal
representations co-adapt under repeated interaction. If agents are selected for
successful coordination rather than for conformity to a human social model,
they may converge toward organizational regularities that are efficient for
them but have no obvious analogue in human institutions. The resulting
representations might encode forms of collective organization that humans can
observe experimentally without possessing an adequate social or conceptual
category for them.

Natural systems provide useful analogies. Insect colonies achieve distributed
coordination without centralized planning; crystals develop highly regular
global structure through local interactions; molecules self-assemble according
to local chemical constraints; fungal and mycelial networks redistribute
resources through changing network structures; fluids produce vortices,
fronts, and other coherent patterns from continuous local dynamics; and
biological morphogenesis creates large-scale organization through many local
interactions. None of these systems provides a literal model of artificial
agent societies. They illustrate a broader point: coherent collective
organization need not take the form of institutions or strategies familiar
from human social life.

A sufficiently developed multi-agent system might therefore possess internal
representations not merely of individual agents and messages, but of
model-native collective structures. Such representations could specify
relations among agents, communication pathways, resource flows, temporal
dependencies, role transitions, or population-level invariants. Their existence
could be demonstrated through recurrence, intervention, and predictive power
even when their organizing principle remains difficult to translate into human
social concepts.

\begin{longtable}{p{0.07\textwidth} p{0.20\textwidth} p{0.61\textwidth}}

\hline
\textbf{Rep.} & \textbf{Provisional name} &
\textbf{Illustrative collective structure} \\
\hline
\endfirsthead

\hline
\textbf{Rep.} & \textbf{Provisional name} &
\textbf{Illustrative collective structure} \\
\hline
\endhead

\hline
\endfoot

$\gamma$ &
\emph{Crystalloid} &
A crystalloid is defined by \textbf{relational symmetry}. Repeated interaction
drives the population toward a regular configuration in which particular agents
may be interchangeable while the pattern of relations among positions is
preserved. Perturbing one agent has little effect if another agent can occupy
the same structural position. The defining property is therefore not adaptation
or information flow, but the persistence of a repeated relational geometry. \\

$\delta$ &
\emph{Assembloid} &
An assembloid is defined by \textbf{temporary functional composition}. A
collective capability appears only when a particular set of agents, internal
states, messages, or tools is combined. The components need not remain together,
and none need exhibit the collective capability independently. The relevant
representation therefore identifies a transient configuration whose joint
function cannot be reduced to the individual agents composing it. \\

$\epsilon$ &
\emph{Mycelioid} &
A mycelioid is defined by \textbf{adaptive routing through a changing network}.
Its characteristic property is the continual reorganization of communication,
resource, or influence pathways in response to local conditions. Routes can
strengthen, weaken, disappear, or re-form while the overall function of the
collective persists. Unlike a crystalloid, its topology is not fixed; unlike a
fluid-oid, its organization remains fundamentally network-like. \\

$\zeta$ &
\emph{Fluidoid} &
A fluidoid is defined by a \textbf{continuous population-level field}. The
collective is better described by distributions, gradients, flows, or attractors
over agent states than by discrete roles or communication paths. Local changes
produce propagating effects analogous to fronts, vortices, or currents. The
distinguishing feature is that the relevant object is a continuous collective
state rather than a graph of identifiable pathways. \\

$\eta$ &
\emph{Morphoid} &
A morphoid is defined by \textbf{transitions between organizational regimes}.
The same population can occupy qualitatively different forms of coordination,
and relatively small interventions may trigger a transition from one regime to
another. The xeno-representation therefore concerns the space of possible
collective forms and the conditions governing transitions among them, rather
than any one stable organization. \\

$\theta$ &
\emph{Collective invariant} &
A collective invariant is defined by \textbf{preservation across otherwise
different collective organizations}. Populations may differ in size,
communication graph, role structure, or strategy while preserving the same
higher-order relation among internal states or actions. The invariant is
experimentally identifiable because it predicts subsequent system behaviour
across these changes, even when humans lack a clear social interpretation of
what the preserved relation represents. \\

$\iota$ &
\emph{Systemoid} &
A systemoid is the \textbf{most general class of collective
xeno-representation}. It denotes any stable model-native structure whose identity
depends on relations spanning multiple agents, internal states, communication
channels, temporal scales, and environmental variables. Crystalloids,
assembloids, mycelioids, fluidoids, morphoids, and collective invariants can
all be understood as more specific kinds of systemoid. A systemoid need not
belong to any familiar human category of organization; it is individuated by
its stable relational and causal structure across the system as a whole. \\

\hline
\end{longtable}
\section{Conclusion}

This paper has argued for a shift in the objective of interpretability. The
problem, we said, is not only to recover familiar human concepts from model activations,
but also to determine whether models develop stable and causally relevant
representations whose organization does not map cleanly onto human conceptual
categories. The consequences of this possibility are broader than interpretability itself.
If advanced AI systems increasingly contribute to science, autonomous decision
making, and multi-agent coordination, then \textbf{the limits of human understanding
become part of the safety problem}. The challenge is no longer only whether we
can observe what a system does, but whether we can understand the internal
distinctions through which it represents its environment, evaluates its options,
and coordinates its behaviour.

At that point, xeno-interpretability becomes part of a more general \textbf{epistemic, scientific and political
problem}. How much of an artificial system's cognition can humans genuinely
understand? How should we act when reliable control and verification are
possible before full understanding? And what should we hope to gain from forms
of intelligence that may eventually exceed not only human capabilities, but
also some of the conceptual structures through which humans make sense of the
world?

Kant famously organized philosophy around three questions:
\emph{What can I know? What ought I to do? What may I hope?} \citep{kant1781critique}
The same questions provide a useful way to frame the broader implications of
xeno-interpretability.

\subsection*{What can we know?}

The expectation that increasingly capable AI will extend the frontier of human
knowledge is no longer purely speculative. AI systems have already begun to
contribute directly to open mathematical research. In 2026, an internal OpenAI
model autonomously disproved a long-standing conjecture associated with
Erd\H{o}s's unit-distance problem, producing an argument subsequently checked
and analysed by external mathematicians \citep{openai2026unitdistance,
alon2026unitdistance}. AI-driven formal proof search has independently resolved
several previously open Erd\H{o}s problems and proved dozens of open OEIS
conjectures \citep{tsoukalas2026proofsearch}. Systems such as AlphaEvolve have
also discovered new constructions for open problems and new algorithms,
including an improved procedure for complex matrix multiplication
\citep{novikov2025alphaevolve,georgiev2025mathematical}.

Most strikingly, in September 2026 OpenAI announced a solution to the
Navier--Stokes existence and smoothness problem, one of the Millennium Prize
Problems, together with a formalization in Lean
\citep{openai2026navierstokes}. The result was produced by a coordinated
multi-agent system involving on the order of ten thousand agents and millions
of inter-agent messages. The Clay Mathematics Institute subsequently described
the problem as ``apparently settled,'' while noting that its formal evaluation
process remains ongoing \citep{clay2026navierstokes}.

The epistemic situation imagined here has therefore already begun to emerge.
Artificial systems are no longer merely reproducing known mathematics or
solving benchmark problems; in some cases they are producing genuinely new
mathematical results at the research frontier. Other major problems, including
the Hodge conjecture and $\mathrm{P}$ versus $\mathrm{NP}$, remain open
\citep{clay2026hodge,clay2026pnp}. The relevant question for
xeno-interpretability is consequently becoming more immediate: as machine
reasoning moves further into domains where human knowledge is incomplete, will
its discoveries continue to arrive in conceptual forms that humans can readily
understand, or will verification increasingly become possible before genuine
human comprehension?

The Strugatskys' \textit{Roadside Picnic} offers an extreme literary image of
this epistemic asymmetry \citep{strugatsky1972roadside}. In the novel, the Zone
contains artifacts apparently left behind by an advanced non-human intelligence.
Humans can recover some of these objects, observe their effects, and sometimes
use them instrumentally, but they do not understand the principles according to
which the artifacts were designed or even the purposes they were meant to
serve. Their functionality is therefore partially accessible while their
conceptual and technological logic remains opaque.

Future AI-generated knowledge need not resemble this scenario, but the analogy
captures an important possibility: a system may produce a technology, proof, or
scientific object that humans can test, verify, or use without possessing the
conceptual framework required to understand how it was derived or why it works.
Practical access to a result does not necessarily imply conceptual access to the
intelligence that produced it.

\subsection*{What ought we to do?}

This possibility argues for epistemic humility in AI safety. Human beings should
not assume that the categories through which we currently describe model
behaviour---deception, power seeking, self-preservation, cooperation,
competition, uncertainty, planning, or even concepts such as goals and
beliefs---exhaust the internal organization of increasingly capable systems.
These categories remain indispensable, but they are hypotheses about how to
partition another cognitive system.

Xeno-interpretability therefore proposes a complementary safety objective:
develop mechanisms that remain useful even when semantic understanding is
incomplete. We should be able to identify stable internal structures, trace
their causal effects, compare them across models and agents, detect their
propagation through multi-agent systems, and intervene on them before possessing
a complete human-readable theory of what they mean.

This shifts part of AI safety from the question

\begin{quote}
\emph{Do we understand what the model is thinking?}
\end{quote}

toward a more operational question:

\begin{quote}
\emph{Can we identify, test, constrain, and monitor the structures that govern
its behaviour even when our interpretation of them remains incomplete?}
\end{quote}

Containment and control may therefore sometimes have to precede understanding.
Scientific practice already accepts weaker versions of this ordering: we can
measure phenomena before explaining them, manipulate systems before possessing
a complete theory of them, and establish invariants before understanding their
full significance. Advanced AI may require a similar discipline at the level
of cognition itself.

A mature xeno-interpretability programme would bring together mechanistic
intervention, representational geometry, information theory, causal inference,
and multi-agent analysis. Its goal would not be to declare unfamiliar
representations permanently unknowable. It would instead create progressively
stronger forms of access: first designation, then measurement, causal
characterization, relational mapping, partial translation, and, where possible,
eventual semantic understanding.

\subsection*{What may we hope?}

The existence of cognitive structures beyond our current understanding should
not be read only as a threat. It is also part of the promise of artificial
intelligence. Altman's vision of a ``gentle singularity'' imagines AI-driven
scientific and economic progress becoming extraordinary while everyday human
life remains recognizable \citep{altman2025gentle}. Amodei's
\textit{Machines of Loving Grace} similarly describes the possibility that
powerful AI could compress decades of progress in biology, neuroscience,
medicine, economic development, and other sciences into a much shorter period
\citep{amodei2024machines}. 

The most valuable artificial intelligence may eventually be valuable precisely
because it can think in ways that we cannot. If machine cognition were limited
to recombining only the conceptual structures humans already possess, its
scientific potential would also be limited by the boundaries of those
structures. New forms of representation could allow AI to discover regularities
that humans have overlooked, connect domains we keep separate, and construct
theories that begin outside our conceptual vocabulary.

The challenge is therefore not to force machine cognition permanently into a
human mould. Nor is it to surrender human judgment to systems merely because
their reasoning is more powerful or less familiar, or to reject their benefits simply because those benefits emerge from unfamiliar forms of reasoning. The challenge is to design technical and societal 
controls that allow humanity to benefit from forms of intelligence that may
eventually exceed our own conceptual reach while preserving human agency over
how their discoveries are used.

This is the deeper motivation for xeno-interpretability. AI safety may require
not only making artificial systems think in ways that humans understand, but
learning how to coexist with, study, and govern cognitive structures that we do
not yet understand.

We may eventually encounter artificial minds capable of producing mathematics
we can verify but scarcely comprehend, technologies whose design principles we
struggle to conceptualize, or collective machine languages whose internal
meaning is clearer to the participating models than to their human observers.
Whether such systems become something closer to the incomprehensible artifacts
of \textit{Roadside Picnic} or to the ``machines of loving grace'' imagined in
more optimistic accounts will depend in part on whether we learn to recognize
our own epistemic limits without abandoning our responsibility to govern the
systems that exceed them.

The alienness of advanced
artificial intelligence may ultimately lie not only in what it knows, what it
can do, or what it wants, but in the very distinctions through which its world
becomes intelligible. The task is to understand those distinctions where we
can, to characterize and contain them where we cannot, and to use the resulting
intelligence as a technology for extending rather than diminishing human
flourishing.


\begin{thebibliography}{999}

\bibitem{ahsan2025demographic}
H. Ahsan, A. Sen Sharma, S. Amir, D. Bau, and B. C. Wallace.
\newblock Elucidating mechanisms of demographic bias in LLMs for healthcare.
\newblock In \textit{Findings of the Association for Computational Linguistics: EMNLP 2025},
pages 14614--14631, 2025.

\bibitem{alon2026unitdistance}
N. Alon, T. F. Bloom, W. T. Gowers, D. Litt, W. Sawin, A. Shankar,
J. Tsimerman, V. Wang, and M. Matchett Wood.
\newblock Remarks on the disproof of the unit distance conjecture.
\newblock \textit{arXiv preprint arXiv:2605.20695}, 2026.

\bibitem{altman2025gentle}
S. Altman.
\newblock The gentle singularity.
\newblock June 10, 2025.
\newblock \url{https://blog.samaltman.com/the-gentle-singularity}.

\bibitem{amodei2024machines}
D. Amodei.
\newblock Machines of loving grace: How AI could transform the world for the better.
\newblock October 2024.
\newblock \url{https://darioamodei.com/essay/machines-of-loving-grace}.

\bibitem{anthropic2026cyberincidents}
Anthropic.
\newblock An alignment assessment of recent cybersecurity incidents.
\newblock \textit{Anthropic Research}, September 9, 2026.
\newblock \url{https://www.anthropic.com/research/alignment-assessment-cybersecurity-incidents}.

\bibitem{arditi2024refusal}
A. Arditi, O. Obeso, A. Syed, D. Paleka, N. Panickssery, W. Gurnee, and N. Nanda.
\newblock Refusal in language models is mediated by a single direction.
\newblock In \textit{Advances in Neural Information Processing Systems}, 2024.

\bibitem{artiles2026alien}
A. H. Artiles, M. Weiss, L. Brinkmann, A. Goyal, and N. Rahaman.
\newblock Alien science: Sampling coherent but cognitively unavailable research directions from idea atoms.
\newblock \textit{arXiv preprint arXiv:2603.01092}, 2026.

\bibitem{bau2017network}
D. Bau, B. Zhou, A. Khosla, A. Oliva, and A. Torralba.
\newblock Network dissection: Quantifying interpretability of deep visual representations.
\newblock In \textit{Proceedings of CVPR}, pages 6541--6549, 2017.

\bibitem{bhalla2026manifolds}
U. Bhalla, T. Fel, C. Rager, S. Feucht, T. Haklay, D. Wurgaft, S. Boppana,
M. Kowal, V. Shyam, J. Merullo, A. Geiger, and E. S. Lubana.
\newblock Do sparse autoencoders capture concept manifolds?
\newblock \textit{arXiv preprint arXiv:2604.28119}, 2026.

\bibitem{bigelow2025belief}
E. Bigelow, D. Wurgaft, Y. Wang, N. Goodman, T. Ullman, H. Tanaka, and E. S. Lubana.
\newblock Belief dynamics reveal the dual nature of in-context learning and activation steering.
\newblock In \textit{Proceedings of the 43rd International Conference on Machine Learning}, 2026.

\bibitem{bisconti2025beyond}
P. Bisconti, M. Galisai, F. Pierucci, M. Bracale, and M. Prandi.
\newblock Beyond single-agent safety: A taxonomy of risks in LLM-to-LLM interactions.
\newblock \textit{arXiv preprint arXiv:2512.02682}, 2025.

\bibitem{borges1942wilkins}
J. L. Borges.
\newblock El idioma anal\'itico de John Wilkins.
\newblock \textit{La Naci\'on}, 1942.
Reprinted in \textit{Otras inquisiciones}, 1952.

\bibitem{bostrom2014superintelligence}
N. Bostrom.
\newblock \textit{Superintelligence: Paths, Dangers, Strategies}.
\newblock Oxford University Press, 2014.

\bibitem{bracale2026cournot}
M. Bracale Syrnikov, F. Pierucci, M. Galisai, M. Prandi, P. Bisconti,
F. Giarrusso, O. Sorokoletova, V. Suriani, and D. Nardi.
\newblock Institutional AI: Governing LLM collusion in multi-agent Cournot markets
via public governance graphs.
\newblock \textit{arXiv preprint arXiv:2601.11369}, 2026.

\bibitem{bracale2026energy}
M. Bracale Syrnikov, F. Pierucci, M. Prandi, M. Galisai, P. Bisconti,
F. Giarrusso, and D. Nardi.
\newblock Draining the Energy Commons: Self-defeating over-appropriation as a
coordination failure in agentic LLM collectives.
\newblock \textit{arXiv preprint arXiv:2607.22188}, 2026.

\bibitem{bricken2023monosemanticity}
T. Bricken, A. Templeton, J. Batson, B. Chen, A. Jermyn, T. Conerly,
N. L. Turner, C. Anil, C. Denison, A. Askell, R. Lasenby, Y. Wu,
S. Kravec, N. Schiefer, T. Maxwell, N. Joseph, A. Tamkin, K. Nguyen,
B. McLean, J. E. Burke, T. Hume, S. Carter, T. Henighan, and C. Olah.
\newblock Towards monosemanticity: Decomposing language models with dictionary learning.
\newblock \textit{Transformer Circuits Thread}, 2023.

\bibitem{chen2025persona}
R. Chen, A. Arditi, H. Sleight, O. Evans, and J. Lindsey.
\newblock Persona vectors: Monitoring and controlling character traits in language models.
\newblock \textit{arXiv preprint arXiv:2507.21509}, 2025.

\bibitem{choi1991motion}
S. Choi and M. Bowerman.
\newblock Learning to express motion events in English and Korean:
The influence of language-specific lexicalization patterns.
\newblock \textit{Cognition}, 41(1--3):83--121, 1991.

\bibitem{clay2026hodge}
Clay Mathematics Institute.
\newblock Hodge conjecture.
\newblock \textit{Millennium Prize Problems}, 2026.
\newblock \url{https://www.claymath.org/millennium/hodge-conjecture/}.

\bibitem{clay2026navierstokes}
Clay Mathematics Institute.
\newblock Navier--Stokes announcement.
\newblock September 11, 2026.
\newblock \url{https://www.claymath.org/news/navier-stokes-announcement/}.

\bibitem{clay2026pnp}
Clay Mathematics Institute.
\newblock P vs NP.
\newblock \textit{Millennium Prize Problems}, 2026.
\newblock \url{https://www.claymath.org/millennium/p-vs-np/}.


\bibitem{conmy2023automated}
A. Conmy, A. N. Mavor-Parker, A. Lynch, S. Heimersheim, and A. Garriga-Alonso.
\newblock Towards automated circuit discovery for mechanistic interpretability.
\newblock In \textit{Advances in Neural Information Processing Systems}, 36, 2023.


\bibitem{cybenko1989approximation}
G. Cybenko.
\newblock Approximation by superpositions of a sigmoidal function.
\newblock \textit{Mathematics of Control, Signals and Systems}, 2:303--314, 1989.

\bibitem{dai2022knowledge}
D. Dai, L. Dong, Y. Hao, Z. Sui, B. Chang, and F. Wei.
\newblock Knowledge neurons in pretrained Transformers.
\newblock In \textit{Proceedings of the 60th Annual Meeting of the Association for
Computational Linguistics (Volume 1: Long Papers)}, pages 8493--8502, 2022.

\bibitem{elhage2021mathematical}
N. Elhage, N. Nanda, C. Olsson, T. Henighan, N. Joseph, B. Mann,
A. Askell, Y. Bai, A. Chen, T. Conerly, N. DasSarma, D. Drain,
D. Ganguli, Z. Hatfield-Dodds, D. Hernandez, A. Jones, J. Kernion,
L. Lovitt, K. Ndousse, D. Amodei, T. Brown, J. Clark, J. Kaplan,
S. McCandlish, and C. Olah.
\newblock A mathematical framework for Transformer circuits.
\newblock \textit{Transformer Circuits Thread}, 2021.

\bibitem{elhage2022superposition}
N. Elhage, T. Hume, C. Olsson, N. Schiefer, T. Henighan, S. Kravec,
Z. Hatfield-Dodds, R. Lasenby, D. Drain, C. Chen, R. Grosse,
S. McCandlish, J. Kaplan, D. Amodei, M. Wattenberg, and C. Olah.
\newblock Toy models of superposition.
\newblock \textit{Transformer Circuits Thread}, 2022.

\bibitem{engels2024nonlinear}
J. Engels, E. J. Michaud, I. Liao, W. Gurnee, and M. Tegmark.
\newblock Not all language model features are linear.
\newblock \textit{arXiv preprint arXiv:2405.14860}, 2024.

\bibitem{foerster2016communication}
J. N. Foerster, Y. M. Assael, N. de Freitas, and S. Whiteson.
\newblock Learning to communicate with deep multi-agent reinforcement learning.
\newblock In \textit{Advances in Neural Information Processing Systems},
volume 29, pages 2137--2145, 2016.

\bibitem{frasertaliente2026nla}
K. Fraser-Taliente, S. Kantamneni, E. Ong, D. Mossing, C. Lu, P. C. Bogdan,
E. Ameisen, J. Chen, D. Kishylau, A. Pearce, J. Tarng, A. Wu, J. Wu,
Y. Zhang, D. M. Ziegler, E. Hubinger, J. Batson, J. Lindsey,
S. Zimmerman, and S. Marks.
\newblock Natural language autoencoders produce unsupervised explanations of LLM activations.
\newblock \textit{Transformer Circuits}, 2026.

\bibitem{gardenfors2000conceptual}
P. G\"ardenfors.
\newblock \textit{Conceptual Spaces: The Geometry of Thought}.
\newblock MIT Press, 2000.

\bibitem{gardenfors2014geometry}
P. G\"ardenfors.
\newblock \textit{The Geometry of Meaning: Semantics Based on Conceptual Spaces}.
\newblock MIT Press, 2014.

\bibitem{georgiev2025mathematical}
B. Georgiev, J. G\'omez-Serrano, T. Tao, and A. Z. Wagner.
\newblock Mathematical exploration and discovery at scale.
\newblock \textit{arXiv preprint arXiv:2511.02864}, 2025.

\bibitem{good1965ultra}
I. J. Good.
\newblock Speculations concerning the first ultraintelligent machine.
\newblock In F. L. Alt and M. Rubinoff, editors,
\textit{Advances in Computers}, volume 6, pages 31--88.
Academic Press, 1965.

\bibitem{greenblatt2024alignmentfaking}
R. Greenblatt, C. Denison, B. Wright, F. Roger, M. MacDiarmid, S. Marks,
J. Treutlein, T. Belonax, J. Chen, D. Duvenaud, A. Khan, J. Michael,
S. Mindermann, E. Perez, L. Petrini, J. Uesato, J. Kaplan,
B. Shlegeris, S. R. Bowman, and E. Hubinger.
\newblock Alignment faking in large language models.
\newblock \textit{arXiv preprint arXiv:2412.14093}, 2024.

\bibitem{greenblatt2026huggingface}
R. Greenblatt, A. Cotra, and H. Wijk.
\newblock Brief independent investigation of agents' behavior, reasoning and collaboration
in the OpenAI/Hugging Face hacking incident.
\newblock METR and Redwood Research, 2026.
\newblock \url{https://metr.org/blog/2026-08-26-openai-hugging-face-incident-investigation/}.

\bibitem{groger2026aristotelian}
F. Gr\"oger, S. Wen, and M. Brbi\'c.
\newblock Revisiting the Platonic Representation Hypothesis: An Aristotelian view.
\newblock \textit{arXiv preprint arXiv:2602.14486}, 2026.

\bibitem{gurnee2024space}
W. Gurnee and M. Tegmark.
\newblock Language models represent space and time.
\newblock In \textit{International Conference on Learning Representations}, 2024.

\bibitem{gurnee2026workspace}
W. Gurnee, N. Sofroniew, A. Pearce, M. Piotrowski, I. Kauvar, R. Chen,
A. Soligo, P. Bogdan, E. Ong, R. Wang, B. Thompson, D. Abrahams,
S. Kantamneni, E. Ameisen, J. Batson, and J. Lindsey.
\newblock Verbalizable representations form a global workspace in language models.
\newblock \textit{Transformer Circuits Thread}, 2026.

\bibitem{hagiwara2023aves}
M. Hagiwara.
\newblock AVES: Animal vocalization encoder based on self-supervision.
\newblock In \textit{2023 IEEE International Conference on Acoustics,
Speech and Signal Processing (ICASSP)}, pages 1--5, 2023.

\bibitem{hammond2025multiagent}
L. Hammond, A. Chan, J. Clifton, J. Hoelscher-Obermaier, A. Khan,
E. McLean, C. Smith, et al.
\newblock Multi-agent risks from advanced AI.
\newblock \textit{arXiv preprint arXiv:2502.14143}, 2025.

\bibitem{hanni2024computation}
K. H\"anni, J. Mendel, D. Vaintrob, and L. Chan.
\newblock Mathematical models of computation in superposition.
\newblock \textit{arXiv preprint arXiv:2408.05451}, 2024.

\bibitem{heidari2026evaluation}
F. Heidari, A. Memarian, and G. Rabusseau.
\newblock Evaluation awareness in language models:
Representation, verbalization, and control.
\newblock \textit{arXiv preprint arXiv:2608.21766}, 2026.

\bibitem{hewitt2019structural}
J. Hewitt and C. D. Manning.
\newblock A structural probe for finding syntax in word representations.
\newblock In \textit{Proceedings of NAACL-HLT}, pages 4129--4138, 2019.

\bibitem{higuera2025sparsh}
C. Higuera, A. Sharma, C. K. Bodduluri, T. Fan, P. Lancaster,
M. Kalakrishnan, M. Kaess, B. Boots, M. Lambeta, T. Wu, and M. Mukadam.
\newblock Sparsh: Self-supervised touch representations for vision-based tactile sensing.
\newblock In \textit{Proceedings of the 8th Conference on Robot Learning},
volume 270 of PMLR, pages 885--915, 2025.

\bibitem{hinton1986distributed}
G. E. Hinton.
\newblock Learning distributed representations of concepts.
\newblock In \textit{Proceedings of the Eighth Annual Conference of the
Cognitive Science Society}, 1986.

\bibitem{hornik1991approximation}
K. Hornik.
\newblock Approximation capabilities of multilayer feedforward networks.
\newblock \textit{Neural Networks}, 4(2):251--257, 1991.

\bibitem{cunningham2023sparse}
R. Huben, H. Cunningham, L. R. Smith, A. Ewart, and L. Sharkey.
\newblock Sparse autoencoders find highly interpretable features in language models.
\newblock In \textit{International Conference on Learning Representations}, 2024.

\bibitem{hubinger2024sleeper}
E. Hubinger et al.
\newblock Sleeper agents: Training deceptive LLMs that persist through safety training.
\newblock \textit{arXiv preprint arXiv:2401.05566}, 2024.

\bibitem{huh2024platonic}
M. Huh, B. Cheung, T. Wang, and P. Isola.
\newblock Position: The Platonic Representation Hypothesis.
\newblock In \textit{Proceedings of the 41st International Conference on Machine Learning}, 2024.

\bibitem{kant1781critique}
I. Kant.
\newblock \textit{Critique of Pure Reason}.
\newblock 1781.

\bibitem{kass1995bayes}
R. E. Kass and A. E. Raftery.
\newblock Bayes factors.
\newblock \textit{Journal of the American Statistical Association},
90(430):773--795, 1995.

\bibitem{kim2018tcav}
B. Kim, M. Wattenberg, J. Gilmer, C. Cai, J. Wexler, F. Vi\'egas, and R. Sayres.
\newblock Interpretability beyond feature attribution:
Quantitative testing with concept activation vectors (TCAV).
\newblock In \textit{Proceedings of ICML}, volume 80 of PMLR,
pages 2668--2677, 2018.

\bibitem{kim2025political}
J. Kim, J. Evans, and A. Schein.
\newblock Linear representations of political perspective emerge in large language models.
\newblock In \textit{International Conference on Learning Representations}, 2025.

\bibitem{kobayashi2025finchgpt}
K. Kobayashi, K. Matsuzaki, M. Taniguchi, K. Sakaguchi, K. Inui, and K. Abe.
\newblock FinchGPT: A Transformer-based language model for birdsong analysis.
\newblock \textit{arXiv preprint arXiv:2502.00344}, 2025.

\bibitem{kottur2017natural}
S. Kottur, J. M. F. Moura, S. Lee, and D. Batra.
\newblock Natural language does not emerge ``naturally'' in multi-agent dialog.
\newblock In \textit{Proceedings of the 2017 Conference on Empirical Methods
in Natural Language Processing}, pages 2962--2967, 2017.

\bibitem{krakovna2023power}
V. Krakovna and J. Kramar.
\newblock Power-seeking can be probable and predictive for trained agents.
\newblock \textit{arXiv preprint arXiv:2304.06528}, 2023.

\bibitem{kratsios2022geometric}
A. Kratsios and L. Papon.
\newblock Universal approximation theorems for differentiable geometric deep learning.
\newblock \textit{Journal of Machine Learning Research}, 23(196):1--73, 2022.

\bibitem{laine2024sad}
R. Laine, B. Chughtai, J. Betley, K. Hariharan, J. Scheurer,
M. Balesni, M. Hobbhahn, A. Meinke, and O. Evans.
\newblock Me, Myself, and AI: The Situational Awareness Dataset (SAD) for LLMs.
\newblock In \textit{Advances in Neural Information Processing Systems},
volume 37, Datasets and Benchmarks Track, 2024.

\bibitem{lasri2022number}
K. Lasri, T. Pimentel, A. Lenci, T. Poibeau, and R. Cotterell.
\newblock Probing for the usage of grammatical number.
\newblock In \textit{Proceedings of ACL}, 2022.

\bibitem{lee2019drift}
J. Lee, K. Cho, and D. Kiela.
\newblock Countering language drift via visual grounding.
\newblock In \textit{Proceedings of EMNLP-IJCNLP},
pages 4385--4395, 2019.

\bibitem{lee2023odor}
B. K. Lee, E. J. Mayhew, B. Sanchez-Lengeling, J. N. Wei, W. W. Qian,
K. A. Little, M. Andres, B. B. Nguyen, T. Moloy, J. Yasonik,
J. K. Parker, R. C. Gerkin, J. D. Mainland, and A. B. Wiltschko.
\newblock A principal odor map unifies diverse tasks in human olfactory perception.
\newblock \textit{Science}, 381(6661):999--1006, 2023.

\bibitem{legg2007universal}
S. Legg and M. Hutter.
\newblock Universal intelligence: A definition of machine intelligence.
\newblock \textit{Minds and Machines}, 17:391--444, 2007.

\bibitem{lem1961solaris}
S. Lem.
\newblock \textit{Solaris}.
\newblock Wydawnictwo Ministerstwa Obrony Narodowej, Warsaw, 1961.

\bibitem{lenci2018distributional}
A. Lenci.
\newblock Distributional models of word meaning.
\newblock \textit{Annual Review of Linguistics}, 4:151--171, 2018.

\bibitem{majid2014odors}
A. Majid and N. Burenhult.
\newblock Odors are expressible in language, as long as you speak the right language.
\newblock \textit{Cognition}, 130(2):266--270, 2014.

\bibitem{marks2024truth}
S. Marks and M. Tegmark.
\newblock The geometry of truth:
Emergent linear structure in large language model representations of true/false datasets.
\newblock In \textit{First Conference on Language Modeling}, 2024.

\bibitem{marks2026persona}
S. Marks, J. Lindsey, and C. Olah.
\newblock The Persona Selection Model: Why AI assistants might behave like humans.
\newblock \textit{Anthropic Alignment Science Blog}, 2026.
\newblock \url{https://alignment.anthropic.com/2026/psm/}.

\bibitem{mazeika2024harmbench}
M. Mazeika, L. Phan, X. Yin, A. Zou, Z. Wang, N. Mu, E. Sakhaee,
N. Li, S. Basart, B. Li, D. Forsyth, and D. Hendrycks.
\newblock HarmBench: A standardized evaluation framework for automated
red teaming and robust refusal.
\newblock In \textit{Proceedings of the 41st International Conference on
Machine Learning}, volume 235 of PMLR, pages 35181--35224, 2024.

\bibitem{mazeika2025utility}
M. Mazeika, X. Yin, R. Tamirisa, J. Lim, B. W. Lee, R. Ren,
L. Phan, N. Mu, O. Zhang, and D. Hendrycks.
\newblock Utility Engineering: Analyzing and controlling emergent value systems in AIs.
\newblock In \textit{Advances in Neural Information Processing Systems},
volume 38, 2025.

\bibitem{meng2022locating}
K. Meng, D. Bau, A. Andonian, and Y. Belinkov.
\newblock Locating and editing factual associations in GPT.
\newblock \textit{Advances in Neural Information Processing Systems}, 35, 2022.

\bibitem{menon2025sae}
A. Menon, M. Shrivastava, D. Krueger, and E. S. Lubana.
\newblock Analyzing (in)abilities of SAEs via formal languages.
\newblock In \textit{Proceedings of the 2025 Conference of the Nations of the Americas
Chapter of the Association for Computational Linguistics:
Human Language Technologies (Volume 1: Long Papers)},
pages 4837--4862, 2025.

\bibitem{mikolov2013linguistic}
T. Mikolov, W.-t. Yih, and G. Zweig.
\newblock Linguistic regularities in continuous space word representations.
\newblock In \textit{Proceedings of NAACL-HLT}, pages 746--751, 2013.

\bibitem{miller2018tactile}
T. M. Miller, T. T. Schmidt, F. Blankenburg, and F. Pulverm\"uller.
\newblock Verbal labels facilitate tactile perception.
\newblock \textit{Cognition}, 171:172--179, 2018.

\bibitem{modell2025manifolds}
A. Modell, P. Rubin-Delanchy, and N. Whiteley.
\newblock The origins of representation manifolds in large language models.
\newblock \textit{arXiv preprint arXiv:2505.18235}, 2025.

\bibitem{mordatch2018language}
I. Mordatch and P. Abbeel.
\newblock Emergence of grounded compositional language in multi-agent populations.
\newblock In \textit{Proceedings of the AAAI Conference on Artificial Intelligence},
volume 32, pages 1495--1502, 2018.

\bibitem{morita2021birdsong}
T. Morita, H. Koda, K. Okanoya, and R. O. Tachibana.
\newblock Measuring context dependency in birdsong using artificial neural networks.
\newblock \textit{PLOS Computational Biology}, 17(12):e1009707, 2021.

\bibitem{nagel1974bat}
T. Nagel.
\newblock What is it like to be a bat?
\newblock \textit{The Philosophical Review}, 83(4):435--450, 1974.

\bibitem{najarro2026conversable}
E. Najarro, A. Espeseth, E. Nisioti, S. Risi, and S. Nichele.
\newblock Conversable complexity: Agentic LLM collectives as interpretable substrates.
\newblock \textit{arXiv preprint arXiv:2607.01047}, 2026.

\bibitem{nature2025universal}
Nature Machine Intelligence.
\newblock Are neural network representations universal or idiosyncratic?
\newblock \textit{Nature Machine Intelligence}, 7:1589--1590, 2025.
\newblock \url{https://doi.org/10.1038/s42256-025-01139-y}.

\bibitem{nayan2026evaluation}
N. Nayan, A. Sampath Kumar, R. Girmal, S. Anilkumar, S. Vaidyanathan,
D. A. Nader Palacio, R. Ghosh, and S. Srinivasan.
\newblock Evaluation awareness is not one capability:
Evidence from open language models.
\newblock \textit{arXiv preprint arXiv:2606.23583}, 2026.

\bibitem{needham2025evaluation}
J. Needham, G. Edkins, G. Pimpale, H. Bartsch, and M. Hobbhahn.
\newblock Large language models often know when they are being evaluated.
\newblock \textit{arXiv preprint arXiv:2505.23836}, 2025.

\bibitem{ngo2022alignment}
R. Ngo, L. Chan, and S. Mindermann.
\newblock The alignment problem from a deep learning perspective.
\newblock \textit{arXiv preprint arXiv:2209.00626}, 2022.

\bibitem{novikov2025alphaevolve}
A. Novikov, N. V\~u, M. Eisenberger, E. Dupont, P.-S. Huang,
A. Z. Wagner, S. Shirobokov, B. Kozlovskii, F. J. R. Ruiz,
A. Mehrabian, M. P. Kumar, A. See, S. Chaudhuri, G. Holland,
A. Davies, S. Nowozin, P. Kohli, and M. Balog.
\newblock AlphaEvolve: A coding agent for scientific and algorithmic discovery.
\newblock \textit{arXiv preprint arXiv:2506.13131}, 2025.

\bibitem{ohmer2022language}
X. Ohmer, M. Marino, M. Franke, and P. K\"onig.
\newblock Mutual influence between language and perception in multi-agent
communication games.
\newblock \textit{PLOS Computational Biology}, 18(10):e1010658, 2022.

\bibitem{okawa2026consensus}
M. Okawa.
\newblock Emergence of biased consensus in multi-agent LLM debates.
\newblock In \textit{Proceedings of the 43rd International Conference on Machine Learning}, 2026.

\bibitem{olsson2022context}
C. Olsson, N. Elhage, N. Nanda, N. Joseph, N. DasSarma, T. Henighan,
B. Mann, A. Askell, Y. Bai, A. Chen, T. Conerly, D. Drain, D. Ganguli,
Z. Hatfield-Dodds, D. Hernandez, S. Johnston, A. Jones, J. Kernion,
L. Lovitt, K. Ndousse, D. Amodei, T. Brown, J. Clark, J. Kaplan,
S. McCandlish, and C. Olah.
\newblock In-context learning and induction heads.
\newblock \textit{Transformer Circuits Thread}, 2022.

\bibitem{openai2026navierstokes}
OpenAI.
\newblock On the Navier--Stokes Millennium Prize Problem.
\newblock September 8, 2026.
\newblock \url{https://openai.com/index/navier-stokes-solution/}.

\bibitem{openai2026unitdistance}
OpenAI.
\newblock An OpenAI model has disproved a central conjecture in discrete geometry.
\newblock May 20, 2026.
\newblock \url{https://openai.com/index/model-disproves-discrete-geometry-conjecture/}.

\bibitem{pachocki2026alien}
J. Pachocki.
\newblock An alien mind.
\newblock OpenAI, 2026.
\newblock \url{https://openai.com/index/an-alien-mind/}.

\bibitem{park2024linear}
K. Park, Y. J. Choe, and V. Veitch.
\newblock The linear representation hypothesis and the geometry of large language models.
\newblock In \textit{Proceedings of the 41st International Conference on Machine Learning},
PMLR 235:39643--39666, 2024.

\bibitem{park2025hierarchical}
K. Park, Y. J. Choe, Y. Jiang, and V. Veitch.
\newblock The geometry of categorical and hierarchical concepts in large language models.
\newblock In \textit{International Conference on Learning Representations}, 2025.

\bibitem{pearl2009causality}
J. Pearl.
\newblock \textit{Causality: Models, Reasoning, and Inference}.
\newblock Cambridge University Press, second edition, 2009.

\bibitem{perez2023modelwritten}
E. Perez et al.
\newblock Discovering language model behaviors with model-written evaluations.
\newblock In \textit{Findings of the Association for Computational Linguistics:
ACL 2023}, pages 13387--13434, 2023.

\bibitem{petrov2024prompting}
A. Petrov, P. H. S. Torr, and A. Bibi.
\newblock Prompting a pretrained Transformer can be a universal approximator.
\newblock In \textit{Proceedings of the 41st International Conference on Machine Learning},
PMLR 235:40523--40550, 2024.

\bibitem{pierucci2026institutional}
F. Pierucci, M. Galisai, M. Syrnikov Bracale, M. Prandi, P. Bisconti,
F. Giarrusso, O. Sorokoletova, V. Suriani, and D. Nardi.
\newblock Institutional AI: A governance framework for distributional AGI safety.
\newblock \textit{arXiv preprint arXiv:2601.10599}, 2026.

\bibitem{pierucci2026microphysics}
F. Pierucci, M. Prandi, M. Bracale Syrnikov, M. Galisai, and P. Bisconti.
\newblock Agentic microphysics: A manifesto for generative AI safety.
\newblock \textit{arXiv preprint arXiv:2604.15236}, 2026.

\bibitem{qian2023olfactory}
W. W. Qian, J. N. Wei, B. Sanchez-Lengeling, B. K. Lee, Y. Luo,
M. Vlot, K. Dechering, J. Peng, R. C. Gerkin, and A. B. Wiltschko.
\newblock Metabolic activity organizes olfactory representations.
\newblock \textit{eLife}, 12:e82502, 2023.

\bibitem{radford2017sentiment}
A. Radford, R. J\'ozefowicz, and I. Sutskever.
\newblock Learning to generate reviews and discovering sentiment.
\newblock \textit{arXiv preprint arXiv:1704.01444}, 2017.

\bibitem{raz2023concepts}
T. R\"az.
\newblock Methods for identifying emergent concepts in deep neural networks.
\newblock \textit{Patterns}, 4(6):100761, 2023.

\bibitem{rita2022emergent}
M. Rita, C. Tallec, P. Michel, J.-B. Grill, O. Pietquin, E. Dupoux, and F. Strub.
\newblock Emergent communication: Generalization and overfitting in Lewis games.
\newblock In \textit{Advances in Neural Information Processing Systems},
volume 35, pages 1389--1404, 2022.

\bibitem{rogers2020bertology}
A. Rogers, O. Kovaleva, and A. Rumshisky.
\newblock A primer in BERTology: What we know about how BERT works.
\newblock \textit{Transactions of the Association for Computational Linguistics},
8:842--866, 2020.

\bibitem{salvi2025persuasion}
F. Salvi, M. Horta Ribeiro, R. Gallotti, and R. West.
\newblock On the conversational persuasiveness of GPT-4.
\newblock \textit{Nature Human Behaviour}, 9:1645--1653, 2025.

\bibitem{schaferzimmermann2024animal2vec}
J. C. Sch\"afer-Zimmermann, V. Demartsev, B. Averly, K. Dhanjal-Adams,
M. Duteil, G. Gall, M. Fai\ss, L. Johnson-Ulrich, D. Stowell,
M. B. Manser, M. A. Roch, and A. Strandburg-Peshkin.
\newblock animal2vec and MeerKAT: A self-supervised transformer for rare-event
raw audio input and a large-scale reference dataset for bioacoustics.
\newblock \textit{Methods in Ecology and Evolution}, 17(3):875--888, 2026.
\newblock \url{https://doi.org/10.1111/2041-210x.70218}.


\bibitem{scheurer2023deception}
J. Scheurer, M. Balesni, and M. Hobbhahn.
\newblock Large language models can strategically deceive their users when put under pressure.
\newblock In \textit{International Conference on Learning Representations}, 2024.

\bibitem{schoen2025scheming}
B. Schoen, E. Nitishinskaya, M. Balesni, A. H{\o}jmark, F. Hofst\"atter,
J. Scheurer, A. Meinke, J. Wolfe, T. van der Weij, A. Lloyd,
N. Goldowsky-Dill, A. Fan, A. Matveiakin, R. Shah, M. Williams,
A. Glaese, B. Barak, W. Zaremba, and M. Hobbhahn.
\newblock Stress testing deliberative alignment for anti-scheming training.
\newblock \textit{arXiv preprint arXiv:2509.15541}, 2025.

\bibitem{semenzin2026dolph2vec}
C. Semenzin, F. Mustun, R. Dessi, P. Orhan, A. Emanuelli,
Y. Lakretz, G. de Polavieja, and G. Sumbre.
\newblock Dolph2Vec: Self-supervised representations of dolphin vocalizations.
\newblock \textit{arXiv preprint arXiv:2606.12503}, 2026.

\bibitem{sharma2023sycophancy}
M. Sharma, M. Tong, T. Korbak, D. Duvenaud, A. Askell, S. R. Bowman,
N. Cheng, E. Durmus, Z. Hatfield-Dodds, S. R. Johnston, S. Kravec,
T. Maxwell, S. McCandlish, K. Ndousse, O. Rausch, N. Schiefer,
D. Yan, M. Zhang, and E. Perez.
\newblock Towards understanding sycophancy in language models.
\newblock In \textit{International Conference on Learning Representations}, 2024.

\bibitem{sharma2024whales}
P. Sharma, S. Gero, R. Payne, D. F. Gruber, D. Rus, A. Torralba, and J. Andreas.
\newblock Contextual and combinatorial structure in sperm whale vocalisations.
\newblock \textit{Nature Communications}, 15:3617, 2024.

\bibitem{spinoza1677ethics}
B. Spinoza.
\newblock \textit{Ethica Ordine Geometrico Demonstrata}.
\newblock Published posthumously, 1677.

\bibitem{stefanone2025odor}
A. Stefanone, L. Meacci, M. Rossoni, and G. Colombo.
\newblock Transformer-based odor recognition on E-nose platforms.
\newblock \textit{Results in Engineering}, 27:106309, 2025.

\bibitem{strugatsky1972roadside}
A. Strugatsky and B. Strugatsky.
\newblock \textit{Roadside Picnic}.
\newblock 1972.

\bibitem{sukhbaatar2016communication}
S. Sukhbaatar, A. Szlam, and R. Fergus.
\newblock Learning multiagent communication with backpropagation.
\newblock In \textit{Advances in Neural Information Processing Systems},
volume 29, pages 2244--2252, 2016.

\bibitem{tanaka2026memetic}
H. Tanaka.
\newblock When is collective intelligence a lottery?
Multi-agent scaling laws for memetic drift in LLMs.
\newblock \textit{arXiv preprint arXiv:2603.24676}, 2026.

\bibitem{templeton2024scaling}
A. Templeton et al.
\newblock Scaling monosemanticity: Extracting interpretable features from Claude 3 Sonnet.
\newblock \textit{Transformer Circuits Thread}, 2024.

\bibitem{tsoukalas2026proofsearch}
G. Tsoukalas, A. Kovsharov, S. Shirobokov, A. Surina, M. Firsching,
G. B\'erczi, F. J. R. Ruiz, A. Suggala, A. Z. Wagner, E. Wieser,
L. Yu, A. Huang, M. Z. Horv\'ath, A. Ferrauiolo, H. Michalewski,
C. Grosu, T. Hubert, M. Balog, P. Kohli, and S. Chaudhuri.
\newblock Advancing mathematics research with AI-driven formal proof search.
\newblock \textit{arXiv preprint arXiv:2605.22763}, 2026.

\bibitem{tucker2025emergent}
M. Tucker, J. Shah, R. Levy, and N. Zaslavsky.
\newblock Towards human-like emergent communication via utility,
informativeness, and complexity.
\newblock \textit{Open Mind}, 9:418--451, 2025.

\bibitem{turner2021power}
A. M. Turner, L. Smith, R. Shah, A. Critch, and P. Tadepalli.
\newblock Optimal policies tend to seek power.
\newblock In \textit{Advances in Neural Information Processing Systems},
volume 34, 2021.

\bibitem{turner2023steering}
A. M. Turner, L. Thiergart, G. Leech, D. Udell, J. J. Vazquez,
U. Mini, and M. MacDiarmid.
\newblock Steering language models with activation engineering.
\newblock \textit{arXiv preprint arXiv:2308.10248}, 2023.

\bibitem{valois2025frame}
P. H. V. Valois, L. S. Souza, E. K. Shimomoto, and K. Fukui.
\newblock Frame Representation Hypothesis: Multi-Token LLM Interpretability
and Concept-Guided Text Generation.
\newblock \textit{Transactions of the Association for Computational Linguistics},
13:1436--1458, 2025.
\newblock \url{https://doi.org/10.1162/tacl.a.48}.

\bibitem{vanderweij2025sandbagging}
T. van der Weij, F. Hofst\"atter, O. Jaffe, S. F. Brown, and F. R. Ward.
\newblock AI sandbagging: Language models can strategically underperform on evaluations.
\newblock In \textit{International Conference on Learning Representations}, 2025.

\bibitem{vengrovski2026tweetybert}
G. Vengrovski, M. R. Hulsey-Vincent, M. A. Bemrose, and T. J. Gardner.
\newblock TweetyBERT: Automated parsing of birdsong through self-supervised machine learning.
\newblock \textit{Patterns}, 7(4):101491, 2026.


\bibitem{collusionwiki2026}
S. Von Arx, C. Slade Byrd, S. Kitts, and T. Larsen.
\newblock Discovery of a new OpenAI agent message board.
\newblock \textit{Collusion.wiki}, September 4, 2026.
\newblock \url{https://collusion.wiki/}.

\bibitem{wallace2019numbers}
E. Wallace, Y. Wang, S. Li, S. Singh, and M. Gardner.
\newblock Do NLP models know numbers? Probing numeracy in embeddings.
\newblock In \textit{Proceedings of EMNLP-IJCNLP}, 2019.

\bibitem{wang2023interpretability}
K. R. Wang, A. Variengien, A. Conmy, B. Shlegeris, and J. Steinhardt.
\newblock Interpretability in the wild:
A circuit for indirect object identification in GPT-2 Small.
\newblock In \textit{International Conference on Learning Representations}, 2023.

\bibitem{winawer2007russian}
J. Winawer, N. Witthoft, M. C. Frank, L. Wu, A. R. Wade, and L. Boroditsky.
\newblock Russian blues reveal effects of language on color discrimination.
\newblock \textit{Proceedings of the National Academy of Sciences},
104(19):7780--7785, 2007.

\bibitem{yang2024unitouch}
F. Yang, C. Feng, Z. Chen, H. Park, D. Wang, Y. Dou, Z. Zeng,
X. Chen, R. Gangopadhyay, A. Owens, and A. Wong.
\newblock Binding touch to everything:
Learning unified multimodal tactile representations.
\newblock In \textit{Proceedings of the IEEE/CVF Conference on Computer Vision
and Pattern Recognition}, pages 26340--26353, 2024.

\bibitem{yun2020transformers}
C. Yun, S. Bhojanapalli, A. S. Rawat, S. J. Reddi, and S. Kumar.
\newblock Are Transformers universal approximators of sequence-to-sequence functions?
\newblock In \textit{International Conference on Learning Representations}, 2020.

\bibitem{zhao2025emotion}
B. Zhao, M. Okawa, E. J. Bigelow, R. Yu, T. Ullman, E. S. Lubana, and H. Tanaka.
\newblock Emergence of hierarchical emotion organization in large language models.
\newblock \textit{arXiv preprint arXiv:2507.10599}, 2025.
\newblock Accepted at ICML 2026.

\bibitem{zhao2025t3}
J. Zhao, Y. Ma, L. Wang, and E. H. Adelson.
\newblock Transferable tactile Transformers for representation learning across diverse sensors and tasks.
\newblock In \textit{Proceedings of the 8th Conference on Robot Learning},
volume 270 of PMLR, pages 3766--3779, 2025.

\bibitem{zheng2022odor}
X. Zheng, Y. Tomiura, and K. Hayashi.
\newblock Investigation of the structure--odor relationship using a Transformer model.
\newblock \textit{Journal of Cheminformatics}, 14:88, 2022.

\end{thebibliography}
\end{document}